\documentclass[11pt]{article}

\usepackage{amsmath}
\usepackage{amssymb}
\usepackage{amsthm}
\usepackage{mathtools}
\usepackage{mathrsfs}

\usepackage{abstract}
\usepackage{appendix}
\usepackage{titlesec}

\usepackage{natbib}
\bibpunct{(}{)}{;}{a}{,}{,}

\usepackage{hyperref}
\hypersetup{
  colorlinks=true,
  linkcolor={blue},
  urlcolor={black},
  citecolor={blue},
  linktoc=all,
}

\usepackage[
  left=1in,
  right=1in,
  top=1in,
  bottom=1in,
  headheight=0pt,
  headsep=5pt
]{geometry}

\usepackage{caption}

\usepackage[shortlabels,inline]{enumitem}

\usepackage{array}
\usepackage{booktabs}

\usepackage{graphicx}
\usepackage{float}

\usepackage{algorithm}
\usepackage[noEnd=false,indLines=false]{algpseudocodex}

\usepackage{xparse}
\usepackage{xstring}

\makeatletter
\newtoks\version
\newtoks\institute
\renewcommand{\@maketitle}{%
  \newpage
  \null
  \vskip 2em%
  \begin{center}%
    {\LARGE\bfseries \@title \par}%
    \vskip 1.5em%
      {\large
        \lineskip .5em%
        \begin{tabular}[t]{c}%
          \@author \\[1ex]
        \end{tabular}\par}%
    \the\institute%
    \vskip 0.5ex%
  \end{center}%
  \par
}
\makeatother

\titleformat{\section}[block]
{\normalfont\LARGE\bfseries}
{\thesection}{1em}{}
\titleformat{\subsection}[block]
{\normalfont\Large\bfseries}
{\thesubsection}{1em}{}
\titleformat{\subsubsection}[block]
{\normalfont\large\bfseries}
{\thesubsubsection}{1em}{}

\AtBeginDocument{
  \setlength{\jot}{8pt}
  \setlength{\abovedisplayskip}{8pt}
  \setlength{\belowdisplayskip}{8pt}
  \setlength{\abovedisplayshortskip}{6pt}
  \setlength{\belowdisplayshortskip}{6pt}

  \RequirePackage[flushmargin]{footmisc}
  \setlength{\footnotesep}{12pt}
}

\theoremstyle{plain}
\newtheorem{theorem}{Theorem}

\newtheorem{lemma}[theorem]{Lemma}
\newtheorem{corollary}[theorem]{Corollary}

\newtheorem{assumption}{Assumption}

\newcommand{\keywords}[1]{\vskip 2ex\par\noindent\normalfont{\bfseries Keywords:} #1}
\newcommand{\email}[1]{\href{mailto:#1}{\nolinkurl{#1}}}

\newcommand{\SECref}[1]{Section~\ref{#1}}
\newcommand{\APPref}[1]{Appendix~\ref{#1}}
\newcommand{\FIGref}[1]{Figure~\ref{#1}}
\newcommand{\TABref}[1]{Table~\ref{#1}}
\newcommand{\THMref}[1]{Theorem~\ref{#1}}

\newcommand{\LEMref}[1]{Lemma~\ref{#1}}
\newcommand{\CORref}[1]{Corollary~\ref{#1}}

\newcommand{\ASMref}[1]{Assumption~\ref{#1}}

\newenvironment{proofof}[1]{
  \begin{proof}[\normalfont\textbf{Proof of #1}]
    }{
  \end{proof}
}

\makeatletter

\def\defprefix@UPPER{{A}{A},{B}{B},{C}{C},{D}{D},{E}{E},{F}{F},{G}{G},{H}{H},{I}{I},{J}{J},{K}{K},{L}{L},{M}{M},{N}{N},{O}{O},{P}{P},{Q}{Q},{R}{R},{S}{S},{T}{T},{U}{U},{V}{V},{W}{W},{X}{X},{Y}{Y},{Z}{Z}}
\def\defprefix@LOWER{{a}{a},{b}{b},{c}{c},{d}{d},{e}{e},{f}{f},{g}{g},{h}{h},{i}{i},{j}{j},{k}{k},{l}{l},{m}{m},{n}{n},{o}{o},{p}{p},{q}{q},{r}{r},{s}{s},{t}{t},{u}{u},{v}{v},{w}{w},{x}{x},{y}{y},{z}{z}}
\def\defprefix@GREEK{%
  {alpha}{\alpha},{beta}{\beta},{gamma}{\gamma},{delta}{\delta},%
  {epsilon}{\epsilon},{varepsilon}{\varepsilon},{zeta}{\zeta},%
  {eta}{\eta},{theta}{\theta},{vartheta}{\vartheta},{iota}{\iota},%
  {kappa}{\kappa},{varkappa}{\varkappa},{lambda}{\lambda},{mu}{\mu},%
  {nu}{\nu},{xi}{\xi},{pi}{\pi},{varpi}{\varpi},{rho}{\rho},%
  {varrho}{\varrho},{sigma}{\sigma},{varsigma}{\varsigma},{tau}{\tau},%
  {upsilon}{\upsilon},{phi}{\phi},{varphi}{\varphi},{chi}{\chi},%
  {psi}{\psi},{omega}{\omega},%
  {Alpha}{A},{Beta}{B},{Gamma}{\Gamma},{Delta}{\Delta},%
  {Epsilon}{E},{Varepsilon}{\varepsilon},{Zeta}{Z},{Eta}{H},%
  {Theta}{\Theta},{Vartheta}{\vartheta},{Iota}{I},{Kappa}{K},%
  {Varkappa}{\varkappa},{Lambda}{\Lambda},{Mu}{M},{Nu}{N},%
  {Xi}{\Xi},{Omicron}{O},{Pi}{\Pi},{Varpi}{\varpi},{Rho}{P},%
  {Varrho}{\varrho},{Sigma}{\Sigma},{Varsigma}{\varsigma},{Tau}{T},%
  {Upsilon}{\Upsilon},{Phi}{\Phi},{Varphi}{\varphi},{Chi}{X},%
  {Psi}{\Psi},{Omega}{\Omega}}
\def\defprefix@define#1#2\@nil#3#4{%
  \expandafter\newcommand\csname #3#1\endcsname{#4{#2}}%
}
\newcommand{\defprefix}[3]{%
  \def\defprefix@list{}%
  \@for\defprefix@alphabet:=#3\do{%
    \expandafter\edef\expandafter\defprefix@list\expandafter{%
      \defprefix@list,\csname defprefix@\defprefix@alphabet\endcsname}%
  }%
  \@for\defprefix@entry:=\defprefix@list\do{%
    \ifx\defprefix@entry\@empty\else
      \expandafter\defprefix@define\defprefix@entry\@nil{#1}{#2}%
    \fi
  }%
  \let\defprefix@alphabet\relax
  \let\defprefix@entry\relax
  \let\defprefix@list\relax
}

\defprefix{bb}{\mathbb}{UPPER}      
\defprefix{cal}{\mathcal}{UPPER}    
\defprefix{frak}{\mathfrak}{UPPER}  
\defprefix{scr}{\mathscr}{UPPER}    
\defprefix{sf}{\mathsf}{UPPER}      

\defprefix{bs}{\boldsymbol}{UPPER,LOWER,GREEK}

\@ifundefined{Dvert}{\DeclarePairedDelimiter\Dvert{\lvert}{\rvert}}{}
\@ifundefined{DVert}{\DeclarePairedDelimiter\DVert{\lVert}{\rVert}}{}
\@ifundefined{Dangle}{\DeclarePairedDelimiter\Dangle{\langle}{\rangle}}{}
\@ifundefined{Dbrace}{\DeclarePairedDelimiter\Dbrace{\lbrace}{\rbrace}}{}
\@ifundefined{Dparen}{\DeclarePairedDelimiter\Dparen{(}{)}}{}
\@ifundefined{Dbrack}{\DeclarePairedDelimiter\Dbrack{[}{]}}{}

\newcommand{\abs}[1]{\if@display\Dvert*{#1}\else\Dvert{#1}\fi}
\newcommand{\norm}[1]{\if@display\DVert*{#1}\else\DVert{#1}\fi}
\newcommand{\inner}[1]{\if@display\Dangle*{#1}\else\Dangle{#1}\fi}
\newcommand{\family}[1]{\if@display\Dbrace*{#1}\else\Dbrace{#1}\fi}
\newcommand{\Tr}[1]{\if@display\operatorname{Tr} \Dparen*{#1}\else\operatorname{Tr} \Dparen{#1}\fi}
\newcommand{\Ran}[1]{\if@display\operatorname{Ran} \Dparen*{#1}\else\operatorname{Ran} \Dparen{#1}\fi}

\ProvideDocumentCommand{\EE}{o m}{%
  \if@display
    \IfNoValueTF{#1}{\operatorname{\mathbb{E}} \Dbrack*{#2}}{\operatorname{\mathbb{E}}_{#1} \Dbrack*{#2}}%
  \else
    \IfNoValueTF{#1}{\operatorname{\mathbb{E}} \Dbrack{#2}}{\operatorname{\mathbb{E}}_{#1} \Dbrack{#2}}%
  \fi}
\ProvideDocumentCommand{\PP}{o m}{%
  \if@display
    \IfNoValueTF{#1}{\mathbb{P} \Dbrace*{#2}}{\mathbb{P}_{#1} \Dbrace*{#2}}%
  \else
    \IfNoValueTF{#1}{\mathbb{P} \Dbrace{#2}}{\mathbb{P}_{#1} \Dbrace{#2}}%
  \fi}

\newcommand{\biggg}{\bBigg@{3}}
\newcommand{\Biggg}{\bBigg@{3.5}}

\makeatother

\newcommand{\dd}{\mathrm{d}}
\newcommand{\ee}{\mathrm{e}}
\newcommand{\eqspace}{\hphantom{{}={}}}  

\newcommand{\MID}{\,\middle|\,}

\newcommand{\ind}[1]{\mathbf{1}_{ #1 }}

\newcommand{\ROM}[1]{\uppercase\expandafter{\romannumeral #1}}

\let\oldforall\forall
\renewcommand{\forall}{\mathrel{\oldforall}}
\let\oldexists\exists
\renewcommand{\exists}{\mathrel{\oldexists}}

\newcommand{\rstar}{r^{\star}}
\newcommand{\rhat}{\hat{r}}

\newcommand{\fstar}{f^{\star}}
\newcommand{\fhat}{\hat{f}}
\newcommand{\fapp}{\fhat_{B_{N}}^{\,\mathrm{app}}}
\newcommand{\fref}{\fhat_{B_{N}}^{\,\mathrm{ref}}}
\newcommand{\hhat}{\hat{h}}

\newcommand{\TN}{\operatorname{Clip}_{B_{N}}}

\newcommand{\Risk}[1]{\if@display\calR \Dbrack*{#1}\else\calR \Dbrack{#1}\fi}
\newcommand{\RiskN}[1]{\if@display\widehat{\calR}_{N} \Dbrack*{#1}\else\widehat{\calR}_{N} \Dbrack{#1}\fi}
\newcommand{\RiskNv}[1]{\if@display\widehat{\calR}_{v} \Dbrack*{#1}\else\widehat{\calR}_{v} \Dbrack{#1}\fi}
\newcommand{\ExRisk}[1]{\if@display\calE \Dbrack*{#1}\else\calE \Dbrack{#1}\fi}

\newcommand{\TreeClass}{\calF_{K,L}}
\newcommand{\ClassClip}{\TreeClass(B_{N})}
\newcommand{\Split}{\widehat{\calC}_{j}}

\newcommand{\phihat}{\hat{\phi}}
\newcommand{\ahat}{\hat{a}}
\newcommand{\bhat}{\hat{b}}
\newcommand{\Ghat}{\widehat{G}}
\newcommand{\Hhat}{\widehat{H}}
\newcommand{\Pihat}{\widehat{\Pi}}
\newcommand{\thetahat}{\hat{\theta}}
\newcommand{\Dhat}{\widehat{\calD}}
\newcommand{\mhat}{\widehat{m}}

\newcommand{\kratio}{\kappa_{\mathrm{r}}}
\newcommand{\kinput}{\kappa_{\mathrm{in}}}
\newcommand{\alphabar}{\bar{\alpha}}

\newcommand{\Resolution}{\widehat{\Delta}_{j}}
\newcommand{\PartionApp}{\Pihat^{\,\mathrm{app}}}
\newcommand{\Opterror}{\hat{\varepsilon}_{K, L}}

\newcommand{\Pdim}{\operatorname{Pdim}}

\numberwithin{equation}{section}

\graphicspath{{figures/}}

\begin{document}

\title{Hellinger Decision Trees for Density Ratio\\Estimation: Finite-Sample Guarantees}

\author{Ren-Rui Liu\( ^{1} \) \and Zheng-Chu Guo\( ^{1, \ast} \)}

\footnotetext[1]{School of Mathematical Sciences, Zhejiang University, Hangzhou 310058, China}
\renewcommand{\thefootnote}{\fnsymbol{footnote}}
\footnotetext[1]{This work was supported by the National Natural Science Foundation of China (Project No.\ 12271473). The corresponding author is Zheng-Chu Guo. Email address: \href{mailto:guozc@zju.edu.cn}{guozc@zju.edu.cn}.}
\renewcommand{\thefootnote}{\arabic{footnote}}

\date{}

\maketitle

\begin{abstract}
  Density ratios play a central role in two-sample comparison, distribution-shift adaptation, and generative-model evaluation. However, estimating a density ratio through separate estimation of the underlying densities can be unnecessarily challenging. In this paper, we develop a direct approach to the log density ratio estimation by minimizing a balanced Hellinger risk over additive ensembles of axis-aligned decision trees, trained using Newton boosting. We establish finite-sample guarantees for the estimator produced by the algorithm and derive an explicit convergence rate. Under suitable tail and H\"{o}lder smoothness conditions, together with an additional assumption on the progress made at each boosting iteration, our rate matches, up to logarithmic factors, the classical minimax-optimal \( \scrL^{2} \) rate for function estimation. A key technical contribution is an optimization analysis that exploits the special exponential structure of the Hellinger risk. In particular, we show that the decrease in the second-order Taylor approximation (which is computable) of the empirical risk controls the actual decrease in that risk, yielding a direct optimization guarantee for the boosting algorithm. This allows us to establish statistical convergence guarantees for the algorithm's output without relying on an idealized empirical-risk minimizer. Experiments on synthetic data corroborate the predicted convergence behavior, while experiments on public datasets demonstrate the practical effectiveness of the proposed estimator.

  \keywords{Density ratio estimation, Hellinger risk, additive trees, Newton boosting, finite-sample analysis}
\end{abstract}

\section{Introduction}
\label{sec: (sec-1) introduction}

The density ratio \( \rstar = \dd P / \dd Q \) describes the relative mass assigned by two distributions \( P \) and \( Q \) at each point in the input space. It converts expectations under \( Q \) into expectations under \( P \), and appears in various areas including two-sample comparison, change-point detection, distribution-shift correction, divergence estimation, and generative-model evaluation. Estimating the two densities \emph{separately} and then taking their ratio requires recovery of two high-dimensional functions and may amplify errors where the estimated denominator is small. Instead, \emph{direct} density ratio estimation minimizes a risk whose population optimizer is \( \rstar \) (or a prescribed transformation of it, such as \( \log \rstar \)), and estimates only the target function required by the task.

Many direct density ratio estimators fit a function by minimizing an empirical risk computed from the two samples \citep{Sugiyama2012DensityRE}. Kernel-based methods, for example, represent the ratio using kernel functions and estimate their coefficients from the observations \citep{Kanamori2009LeastSquaresAD}. Direct estimation is also connected to binary classification. If observations from \( P \) and \( Q \) are assigned different class labels with equal prior probabilities, the density ratio equals the posterior probability of the \( P \) label divided by that of the \( Q \) label \citep{Qin1998InferencesCS}. This connection provides a way to construct density ratio estimators using classification models, with the corresponding estimation risk determined by the classification loss \citep{Zellinger2025BinaryLD}.

Boosted trees are widely used for binary classification and can therefore also be used to estimate density ratios. A boosted tree model represents its prediction function as a sum of trees, each assigning constant values to regions of the input space. These trees are fitted sequentially, with each new tree added to reduce the empirical risk of the current model. In Newton boosting, the new tree is fitted using a second-order approximation to this risk, as in XGBoost \citep{Chen2016XGBoostST} and LightGBM \citep{Ke2017LightGBMHE}.

This paper studies direct log density ratio estimation under the \emph{balanced Hellinger risk} (formally defined in \eqref{eq: (sec-2) balanced-hellinger-risk}). We use Newton boosting over additive axis-aligned trees to estimate \( \log \rstar \). Both ingredients have appeared in earlier work \citep{Hushchyn2021GeneralizationCD, Awaya2026TwoSampleCA}. Those studies establish balanced Hellinger risk and boosted trees as viable tools for direct density ratio estimation. However, they leave open the finite-sample convergence of the tree estimator returned by the training algorithm.

Establishing such a convergence result requires overcoming three obstacles. First, both the tree structures and the values assigned to their regions depend on the two training samples. The estimation-error bound must therefore hold uniformly over the class of functions that the algorithm can construct. Second, the density ratio and its reciprocal can be unbounded, since one density may become arbitrarily small relative to the other even when the distributions share the same support. Requiring uniform bounds can therefore exclude relevant settings, motivating recent work on estimation with unbounded density ratios \citep{Xu2025EstimatingUD, Zheng2026ErrorAD, Liu2026UnboundedDR}. Approximating the log density ratio by a tree function with bounded values then requires controlling the error contributed by regions where the target takes large positive or negative values. Third, the algorithm fits trees sequentially and stops after finitely many iterations, so the returned estimator need not minimize the empirical risk over the tree class. While \citet{Xu2026StatisticalOO} study the statistical accuracy of trees that minimize empirical risk in regression and classification, a guarantee for the model actually returned by training must also retain the remaining optimization error.

\noindent \textbf{Contributions.} Our analysis addresses these three challenges through a unified error decomposition. Specifically, we decompose the excess risk into estimation, approximation, and optimization errors, establish bounds for each component, and combine them to derive finite-sample generalization guarantees. The main contributions are summarized as follows.
\begin{enumerate}
  \item%
        \textbf{Novel error analysis of the proposed algorithm.} We establish separate bounds for the estimation, approximation, and optimization errors. The estimation error measures the discrepancy between the empirical and population risks, uniformly over the entire class of functions that the algorithm can construct. Its bound is established in \THMref{thm: (sec-3) estimation-error}. The approximation error quantifies the excess risk of the best tree that can be represented by the algorithm, as characterized in \THMref{thm: (sec-3) approximation-error}. The optimization error captures the residual empirical risk after training. A key role in this analysis is played by a property of the balanced Hellinger risk: the second-order decrease, defined in \eqref{eq: (sec-2) second-order-decrease} and greedily maximized by the algorithm, directly controls the decrease in empirical risk, as shown in \THMref{thm: (sec-3) empirical-risk-decrease}. Moreover, under the additional assumption that the second-order decrease captures a fixed fraction of the empirical-risk gap at each iteration (\ASMref{asm: (sec-3) progress-toward-comparison}), we establish an exponential decay of this gap and derive an explicit optimization-error bound in \THMref{thm: (sec-3) optimization-error}.

  \item%
        \textbf{Finite-sample learning guarantees beyond bounded density ratios.} By combining the estimation, approximation, and optimization error bounds, we establish a unified finite-sample excess-risk guarantee in \THMref{thm: (sec-3) unified-finite-sample}. Furthermore, under \ASMref{asm: (sec-3) progress-toward-comparison}, when all \( d \) input coordinates share the same H\"{o}lder smoothness exponent \( \alpha \), \CORref{cor: (sec-3) explicit-rate} yields an excess-risk rate of \( N^{-2 \alpha / (2 \alpha + d)} \), up to logarithmic factors. This agrees with the classical minimax \( \scrL^{2} \) rate for H\"{o}lder-smooth function estimation. We further investigate model selection over a finite candidate set and establish, in \THMref{thm: (sec-3) validation-selection}, a bound on the additional error incurred by using two independent validation samples. Importantly, all these theoretical guarantees accommodate settings in which both the density ratio and its reciprocal may be unbounded.

  \item%
        \textbf{Empirical assessment on synthetic and public datasets.} We conduct extensive experiments on synthetic data and public datasets, as reported in \SECref{sec: (sec-5) experiments}. The experimental results corroborate our theoretical findings and demonstrate the practical effectiveness of the proposed method.
\end{enumerate}

\noindent \textbf{Paper organization.} The remainder of the paper is organized as follows. \SECref{sec: (sec-2) problem-algorithm} introduces the risk functional and the tree class, and describes the proposed training procedure. \SECref{sec: (sec-3) finite-sample} develops the finite-sample error analysis by establishing bounds on the estimation, approximation, and optimization errors, and subsequently deriving a unified finite-sample guarantee. \SECref{sec: (sec-4) related-work} reviews the relevant literature, highlights the connections and differences between our analysis and existing work, and discusses possible extensions. \SECref{sec: (sec-5) experiments} presents the numerical experiments and discusses their results. Finally, \APPref{sec: (app-a) proofs} collects the complete proofs of all theoretical results.

\section{Problem Formulation and Algorithm}
\label{sec: (sec-2) problem-algorithm}

This section defines the balanced Hellinger risk used throughout the paper, introduces the additive axis-aligned tree model, and presents the training procedure.

\subsection{Balanced Hellinger Risk}
\label{subsec: (sec-2) balanced-hellinger-risk}

Let \( P \) and \( Q \) be probability distributions on \( \bbR^{d} \), dominated by a common measure \( \mu \), with respective densities \( p = \dd P / \dd \mu \) and \( q = \dd Q / \dd \mu \). We observe two independent samples,
\[
  X_{1}, \dots, X_{N}
  \overset{\text{i.i.d.}}{\sim} P,
  \quad
  X'_{1}, \dots, X'_{N}
  \overset{\text{i.i.d.}}{\sim} Q.
\]
For simplicity in the subsequent analysis, we take both sample sizes to be \( N \). The target functions are the density ratio and its logarithm:
\begin{equation}
  \rstar(x) = \frac{p(x)}{q(x)},
  \quad
  \fstar(x) = \log \rstar(x).
\end{equation}
For a measurable function \( f \colon \bbR^{d} \to \bbR \), we define the \emph{balanced Hellinger risk} by
\begin{equation}
  \label{eq: (sec-2) balanced-hellinger-risk}
  \Risk{f} =
  \frac{1}{2} \EE[P]{\ee^{-f(X) / 2}} +
  \frac{1}{2} \EE[Q]{\ee^{f(X') / 2}}.
\end{equation}
Its empirical counterpart is
\begin{equation}
  \label{eq: (sec-2) empirical-hellinger-risk}
  \begin{aligned}
    \RiskN{f}
     & =
    \frac{1}{2} \EE[P_{N}]{\ee^{-f(X) / 2}} +
    \frac{1}{2} \EE[Q_{N}]{\ee^{f(X') / 2}}
    \\
     & =
    \frac{1}{2 N} \sum_{i = 1}^{N} \ee^{-f(X_{i}) / 2} +
    \frac{1}{2 N} \sum_{i = 1}^{N} \ee^{f(X'_{i}) / 2},
  \end{aligned}
\end{equation}
where \( P_{N} = \sum_{i = 1}^{N} \delta_{X_{i}} / N \) and \( Q_{N} =  \sum_{i = 1}^{N} \delta_{X'_{i}} / N \) are the empirical distributions of the two samples. The term ``balanced'' reflects the equal weighting of the two expectations in \eqref{eq: (sec-2) balanced-hellinger-risk}, regardless of their respective sample sizes.

The term ``Hellinger'' reflects the connection between the proposed risk and the Hellinger distance. Recall that the squared Hellinger distance \citep{LeCam1986AsymptoticMS} between \( P \) and \( Q \) is defined by
\begin{equation}
  \operatorname{Hel}^{2}(P, Q) =
  \frac{1}{2}
  \int \Dparen*{\sqrt{p} - \sqrt{q}}^{2} \, \dd \mu =
  1 - \int \sqrt{p q} \, \dd \mu.
\end{equation}
Thus, the Hellinger affinity \( \int \sqrt{p q} \, \dd \mu \) equals \( 1 - \operatorname{Hel}^{2}(P, Q) \). To establish the connection between the balanced Hellinger risk \( \Risk{f} \) and the Hellinger affinity \( \int \sqrt{p q} \, \dd \mu \), define
\begin{equation}
  \label{eq: (sec-2) pointwise-risk}
  \Phi_{f}(x) =
  \frac{1}{2} p(x) \ee^{-f(x) / 2} +
  \frac{1}{2} q(x) \ee^{f(x) / 2}.
\end{equation}
Then \( \Risk{f} = \int \Phi_{f}(x) \, \dd \mu(x) \). For any \( x \) such that \( p(x), q(x) > 0 \), the function \( \Phi_{f}(x) \), viewed as a function of \( f(x) \), is strictly convex. Its unique minimizer satisfies
\[
  \ee^{f(x)} = \frac{p(x)}{q(x)} = \rstar(x),
\]
at which the minimum value is \( \sqrt{p(x) q(x)} \). Consequently, the population risk is minimized by \( \fstar = \log \rstar \), yielding
\begin{equation}
  \label{eq: (sec-2) minimum-risk}
  \Risk{\fstar} =
  \int \sqrt{p q} \, \dd \mu.
\end{equation}


We define the corresponding excess risk by
\begin{equation}
  \label{eq: (sec-2) excess-risk}
  \ExRisk{f} =
  \Risk{f} - \Risk{\fstar}.
\end{equation}
A direct calculation yields the equivalent representation
\begin{equation}
  \label{eq: (sec-2) cosh-excess-risk}
  \ExRisk{f} =
  \int \sqrt{p q} \Dparen*{
    \cosh \Dparen*{\frac{f - \fstar}{2}} - 1
  } \dd \mu,
\end{equation}
where \( \cosh(t) = (\ee^{t} + \ee^{-t}) / 2 \) denotes the hyperbolic cosine function. Since \( \cosh(t) \geq 1 \) for all \( t \in \bbR \), this identity immediately implies that \( \ExRisk{f} \geq 0 \). Moreover, it explicitly characterizes the excess risk in terms of the deviation \( f - \fstar \). We use \eqref{eq: (sec-2) cosh-excess-risk} to evaluate the excess risk in the synthetic experiments described in \SECref{subsec: (sec-5) synthetic-experiments}.

\subsection{Hellinger Decision Trees and Newton Boosting}
\label{subsec: (sec-2) axis-aligned-newton-trees}
We begin by describing the structure of a single tree. At each internal node, an \emph{axis-aligned tree} \citep{Tomita2020SparsePO} selects a coordinate \( j \in \family{1, \dots, d} \) and a split point \( c \in \bbR \), partitioning the current region \( A \subseteq \bbR^{d} \) into
\[
  A_{1} =
  A \cap \family{x \MID x^{(j)} \le c},
  \quad
  A_{2} =
  A \cap \family{x \MID x^{(j)} > c},
\]
where \( x^{(j)} \) denotes the \( j \)-th coordinate of \( x \). Recursive splitting yields a partition \( \Pi \) of \( \bbR^{d} \) into disjoint leaf regions, which we call the \emph{leaf partition} of the tree.

A tree assigns a constant value to each leaf. Thus, a tree with at most \( L \) leaves can be represented as
\begin{equation}
  h(x) =
  \sum_{A \in \Pi} \theta(A) \ind{A}(x),
  \quad
  \abs{\Pi} \le L,
\end{equation}
where \( \ind{A} \) denotes the indicator function of \( A \), and \( \theta(A) \) is the value assigned to leaf \( A \).

To make tree construction computationally feasible, we restrict the candidate split points to a finite set. Specifically, for each coordinate \( j \in \family{1, \dots, d} \), we fix a finite set \( \Split \subseteq \bbR \) before training, and require every split along coordinate \( j \) to use a point in \( \Split \). We refer to these points as \emph{allowed split points}. The sets \( \Split \) may depend on the training samples and can be constructed, for example, using candidate split points from the approximate algorithm of XGBoost \citep{Chen2016XGBoostST} or histogram-bin boundaries, as in LightGBM \citep{Ke2017LightGBMHE}. Once constructed, these sets remain fixed throughout the boosting procedure.

The algorithm initializes \( \fhat_{0} = 0 \). At iteration \( t \), it constructs a tree \( \hhat_{t} \) and updates the current estimator according to
\begin{equation}
  \label{eq: (sec-2) boosting-update}
  \fhat_{t + 1} = \fhat_{t} + \nu \hhat_{t},
\end{equation}
where \( 0 < \nu \le 1 \) is the learning rate. After \( K \) iterations, the resulting estimator is
\begin{equation}
  \fhat_{K}(x) =
  \nu \sum_{t = 0}^{K - 1} \hhat_{t}(x).
\end{equation}

Let \( \TreeClass \) denote the class of functions representable as sums of at most \( K \) axis-aligned trees, each having at most \( L \) leaves. To control the range of the estimator, we introduce a clipping threshold \( B_{N} > 0 \), which increases with the sample size \( N \), and define
\begin{equation}
  \label{eq: (sec-2) clipped-estimator}
  \fhat_{K, B_{N}} = \TN \fhat_{K},
  \quad
  \rhat_{K, B_{N}} = \ee^{\fhat_{K, B_{N}}}.
\end{equation}
Here, \( \TN \) denotes the pointwise clipping operator
\begin{equation}
  (\TN f) (x) =
  \max \family{-B_{N}, \min \family{B_{N}, f(x)}}.
\end{equation}
The corresponding clipped function class is
\[
  \ClassClip = \family{\TN f \mid f \in \TreeClass}.
\]
The growing threshold \( B_{N} \) ensures that the clipped estimator remains bounded at each sample size while allowing its range to expand as \( N \) increases, thereby accommodating potentially unbounded log-density ratios.

We next derive the value assigned to each leaf. Fix the current estimator \( \fhat_{t} \) and consider a candidate leaf region \( A \). Adding a constant increment \( z \) to \( \fhat_{t} \) on \( A \) changes the empirical risk through the following leafwise term:
\begin{equation}
  \label{eq: (sec-2) leaf-risk}
  \phihat_{t, A}(z) =
  \frac{1}{2} \ahat_{t}(A) \ee^{-z / 2} +
  \frac{1}{2} \bhat_{t}(A) \ee^{z / 2},
\end{equation}
where
\begin{equation}
  \label{eq: (sec-2) leaf-masses}
  \ahat_{t}(A) =
  \EE[P_{N}]{\ind{A}(X) \ee^{-\fhat_{t}(X) / 2}},
  \quad
  \bhat_{t}(A) =
  \EE[Q_{N}]{\ind{A}(X') \ee^{\fhat_{t}(X') / 2}}.
\end{equation}
These quantities represent the weighted empirical masses of the two samples within \( A \), with weights determined by the current estimator \( \fhat_{t} \).

Evaluating the first and second derivatives of \( \phihat_{t, A} \) at \( z = 0 \) gives
\begin{equation}
  \label{eq: (sec-2) leaf-derivatives}
  \Ghat_{t}(A) =
  \phihat'_{t, A}(0) =
  \frac{\bhat_{t}(A) - \ahat_{t}(A)}{4},
  \quad
  \Hhat_{t}(A) =
  \phihat''_{t, A}(0) =
  \frac{\ahat_{t}(A) + \bhat_{t}(A)}{8}.
\end{equation}
Now fix a candidate leaf partition \( \Pi \) and consider a tree of the form
\[
  h = \sum_{A \in \Pi} \theta(A) \ind{A}.
\]
Since the leaf regions are disjoint and \( h \) is constant on each leaf, the second-order Taylor approximation of the empirical risk around \( \fhat_{t} \) is
\begin{equation}
  \label{eq: (sec-2) second-order-taylor}
  \RiskN{\fhat_{t} + h} \approx
  \RiskN{\fhat_{t}} +
  \sum_{A \in \Pi} \Dparen*{
    \Ghat_{t}(A) \theta(A) +
    \frac{1}{2} \Hhat_{t}(A) \theta(A)^{2}
  }.
\end{equation}
The first term on the right-hand side is independent of \( h \), while the quadratic terms are separable across leaves. Consequently, when \( \ahat_{t}(A) + \bhat_{t}(A) > 0 \), minimizing the quadratic terms with respect to \( \theta(A) \) yields
\begin{equation}
  \label{eq: (sec-2) leaf-value}
  \thetahat_{t}(A) =
  -\frac{\Ghat_{t}(A)}{\Hhat_{t}(A)} =
  2 \frac{
    \ahat_{t}(A) - \bhat_{t}(A)
  }{\ahat_{t}(A) + \bhat_{t}(A)}.
\end{equation}
This is precisely one Newton step for the leafwise objective, starting from \( z = 0 \). If a leaf contains no observations from either sample, we set its value to zero. Since \( \ahat_{t}(A), \bhat_{t}(A) \ge 0 \), the Newton leaf values satisfy \( \abs{\thetahat_{t}(A)} \le 2 \). Constructing boosting trees using such second-order updates is known as \emph{Newton boosting} \citep{Sigrist2022GradientAN}.

Substituting the optimal leaf values \eqref{eq: (sec-2) leaf-value} into the quadratic approximation \eqref{eq: (sec-2) second-order-taylor} gives the predicted reduction in empirical risk associated with a leaf partition \( \Pi \):
\begin{equation}
  \label{eq: (sec-2) second-order-decrease}
  \Dhat_{t}(\Pi) =
  \sum_{A \in \Pi}
  \frac{\Ghat_{t}(A)^{2}}{2 \Hhat_{t}(A)} =
  \sum_{A \in \Pi}
  \frac{
    (\ahat_{t}(A) - \bhat_{t}(A))^{2}
  }{
    4 (\ahat_{t}(A) + \bhat_{t}(A))
  }.
\end{equation}
Throughout, any term with a zero denominator is defined to be zero. We refer to \( \Dhat_{t}(\Pi) \) as the \emph{second-order decrease} associated with \( \Pi \). Although this quantity is derived from a quadratic approximation, it also provides control over the decrease in the original empirical risk, as established in \THMref{thm: (sec-3) empirical-risk-decrease}.

To construct the tree, we use a greedy splitting procedure based on this second-order decrease. Splitting a leaf \( A \) into two child regions \( A_{1} \) and \( A_{2} \) changes \( \Dhat_{t}(\Pi) \) by
\begin{equation}
  \label{eq: (sec-2) split-gain}
  \operatorname{Gain}_{t}(A_{1}, A_{2}) =
  \frac{1}{2} \Dparen*{
    \frac{\Ghat_{t}(A_{1})^{2}}{\Hhat_{t}(A_{1})} +
    \frac{\Ghat_{t}(A_{2})^{2}}{\Hhat_{t}(A_{2})} -
    \frac{\Ghat_{t}(A)^{2}}{\Hhat_{t}(A)}
  }.
\end{equation}
Starting from the root region \( \bbR^{d} \), the algorithm repeatedly selects the candidate split with the largest gain among all current leaves and allowed split points. The procedure terminates when the tree reaches \( L \) leaves or no candidate split yields a positive gain. Let \( \Pihat_{t} \) denote the resulting leaf partition. The tree constructed at iteration \( t \) is then
\begin{equation}
  \label{eq: (sec-2) newton-tree}
  \hhat_{t}(x) =
  \sum_{A \in \Pihat_{t}}
  \thetahat_{t}(A) \ind{A}(x).
\end{equation}
For notational simplicity, we write \( \Dhat_{t} = \Dhat_{t}(\Pihat_{t}) \) for the second-order decrease associated with the constructed tree. The complete training procedure is summarized in Algorithm~\ref{alg: (sec-2) training}. We refer to the resulting estimators as \emph{Hellinger decision trees}, or \emph{Hellinger trees} for short.

The tuning parameters \( K \), \( L \), and \( \nu \) can be selected from a finite set of candidate configurations. For each configuration, we fit a model using the training samples and evaluate its empirical risk on two independent validation samples, one drawn from each distribution. We then select the configuration with the smallest validation risk. The theoretical properties of this selection procedure are established in \THMref{thm: (sec-3) validation-selection}.

\begin{algorithm}[ht]
  \caption{Hellinger decision tree boosting}
  \label{alg: (sec-2) training}
  \begin{algorithmic}[1]
    \Require Training samples \( X_{1}, \dots, X_{N} \sim P \) and \( X'_{1}, \dots, X'_{N} \sim Q \); number of iterations \( K \); leaf limit \( L \); learning rate \( \nu \); clipping threshold \( B_{N} \); allowed split-point sets \( \Split \), \( j = 1, \dots, d \)

    \Ensure Clipped estimators \( \fhat_{K, B_{N}} \) and \( \rhat_{K, B_{N}} \)

    \State Initialize \( \fhat_{0} \leftarrow 0 \).

    \For{\( t = 0, \dots, K - 1 \)}
    \State Initialize \( \Pihat_{t} \leftarrow \family{\bbR^{d}} \).

    \State Compute the weighted empirical masses in \eqref{eq: (sec-2) leaf-masses} for the root region.

    \While{\( \abs{\Pihat_{t}} < L \)}
    \State Evaluate the split gain \eqref{eq: (sec-2) split-gain} for every admissible split of every leaf in \( \Pihat_{t} \).

    \If{no admissible split has positive gain}
    \State \textbf{break}
    \EndIf

    \State Select the split \( (A_{1}, A_{2}) \) with the largest gain.

    \State Replace its parent leaf \( A \) by \( A_{1} \) and \( A_{2} \) in \( \Pihat_{t} \).
    \EndWhile

    \State Compute \( \thetahat_{t}(A) \) for every \( A \in \Pihat_{t} \) using \eqref{eq: (sec-2) leaf-value}.

    \State Construct \( \hhat_{t} \) according to \eqref{eq: (sec-2) newton-tree}.

    \State Update \( \fhat_{t + 1} \leftarrow \fhat_{t} + \nu \hhat_{t} \).
    \EndFor

    \State Set \( \fhat_{K, B_{N}} \leftarrow \TN \fhat_{K} \).

    \State Set \( \rhat_{K, B_{N}} \leftarrow \ee^{\fhat_{K, B_{N}}} \).

    \State \Return \( \fhat_{K, B_{N}} \), \( \rhat_{K, B_{N}} \).
  \end{algorithmic}
\end{algorithm}

\section{Main Results}
\label{sec: (sec-3) finite-sample}

This section establishes finite-sample guarantees for the estimator returned by Algorithm~\ref{alg: (sec-2) training}. We first state the assumptions and decompose the excess risk in \SECref{subsec: (sec-3) assumptions-decomposition}. We then bound the estimation error in \SECref{subsec: (sec-3) estimation-error} (\THMref{thm: (sec-3) estimation-error}) and the approximation error in \SECref{subsec: (sec-3) approximation-error} (\THMref{thm: (sec-3) approximation-error}). Next, \SECref{subsec: (sec-3) optimization-error} relates the second-order decrease to the actual decrease in empirical risk (\THMref{thm: (sec-3) empirical-risk-decrease}) and bounds the remaining optimization error under an additional assumption on the training trajectory (\THMref{thm: (sec-3) optimization-error}). Finally, \SECref{subsec: (sec-3) unified-bound} combines these bounds into a unified excess-risk guarantee (\THMref{thm: (sec-3) unified-finite-sample}), derives an explicit convergence rate (\CORref{cor: (sec-3) explicit-rate}), and controls the additional error from model selection using independent validation samples (\THMref{thm: (sec-3) validation-selection}).

Throughout, the training samples are independent, with observations drawn i.i.d.\ from \( P \) and \( Q \), respectively. The distributions are mutually absolutely continuous, so \( \fstar = \log \rstar \) is finite almost surely but need not be bounded. Unless stated otherwise, \( C \) denotes a positive constant that may vary between occurrences and depend on fixed distributional and smoothness parameters, but not on \( N \) or \( \delta \).

\subsection{Assumptions and Error Decomposition}
\label{subsec: (sec-3) assumptions-decomposition}

We state the first three assumptions here. Assumptions~\ref{asm: (sec-3) density-ratio-tails} and \ref{asm: (sec-3) input-tails} control the tails of the density ratio and the input distributions, while \ASMref{asm: (sec-3) local-smoothness} describes the local smoothness of \( \fstar \). \ASMref{asm: (sec-3) progress-toward-comparison}, which controls the progress made at each iteration, is introduced later, after the relevant quantities have been defined.

\begin{assumption}[Tails of the density ratio]
  \label{asm: (sec-3) density-ratio-tails}
  There is a constant \( \kratio > 0 \) such that
  \[
    \EE[Q]{
      \exp \Dparen*{
        \frac{\rstar(X')}{\kratio}
      }
    } \le 2,
    \quad
    \EE[P]{
      \exp \Dparen*{
        \frac{\rstar(X)^{-1}}{\kratio}
      }
    } \le 2.
  \]
\end{assumption}
The two conditions in \ASMref{asm: (sec-3) density-ratio-tails} give exponentially decaying tails for \( \rstar(X') \) under \( Q \) and \( \rstar(X)^{-1} \) under \( P \), while allowing both quantities to be unbounded. Related moment conditions for unbounded density ratios appear in \citet{Xu2025EstimatingUD}, \citet{Zheng2026ErrorAD}, and \citet{Liu2026UnboundedDR}.

\begin{assumption}[Tails of the input distributions]
  \label{asm: (sec-3) input-tails}
  There is a constant \( \kinput > 0 \) such that
  \[
    \EE[P]{
      \exp \Dparen*{
        \frac{\norm{X}_{\infty}}{\kinput}
      }
    } \le 2,
    \quad
    \EE[Q]{
      \exp \Dparen*{
        \frac{\norm{X'}_{\infty}}{\kinput}
      }
    } \le 2.
  \]
\end{assumption}
The two conditions in \ASMref{asm: (sec-3) input-tails} give exponentially decaying tails for \( \norm{X}_{\infty} \) under \( P \) and \( \norm{X'}_{\infty} \) under \( Q \). This requirement is distinct from \ASMref{asm: (sec-3) density-ratio-tails}: if \( P = Q \), then \( \rstar = 1 \) and \ASMref{asm: (sec-3) density-ratio-tails} holds automatically, whereas the input distribution may still be heavy-tailed.

Before stating the next assumption, we extend \( \fstar = \log \rstar \) from the common support of \( P \) and \( Q \) to \( \bbR^{d} \). With a slight abuse of notation, we continue to denote the extension by \( \fstar \).
\begin{assumption}[Local smoothness]
  \label{asm: (sec-3) local-smoothness}
  There are constants \( 0 < \alpha_{j} \le 1 \) for \( j \in \family{1, \dots, d} \), \( H_{0} > 0 \), and \( \zeta \ge 0 \) such that, for every \( u > 0 \) and every \( x_{1}, x_{2} \in [-u, u]^{d} \),
  \begin{equation}
    \label{eq: (sec-3) local-holder-condition}
    \abs{\fstar(x_{1}) - \fstar(x_{2})} \le
    H_{0} (1 + u)^{\zeta}
    \sum_{j = 1}^{d} \Dparen*{
      \frac{
        \abs{x_{1}^{(j)} - x_{2}^{(j)}}
      }{2 u}
    }^{\alpha_{j}}.
  \end{equation}
\end{assumption}
\ASMref{asm: (sec-3) local-smoothness} states that \( \fstar \) is H\"{o}lder continuous on every cube \( [-u, u]^{d} \). Its H\"{o}lder constant may grow with the scale \( u \), but at most at the polynomial rate \( (1 + u)^{\zeta} \). The exponents \( \alpha_{j} \) allow the smoothness to differ across coordinates. \citet{Xu2025EstimatingUD} use a similar local H\"{o}lder assumption.

Since clipping is \( 1 \)-Lipschitz, \( \TN \fstar \) satisfies \ASMref{asm: (sec-3) local-smoothness} as well. We write
\begin{equation}
  \label{eq: (sec-3) alpha-bar}
  \alphabar^{-1} =
  \sum_{j = 1}^{d} \alpha_{j}^{-1}.
\end{equation}
This quantity will be used to state the approximation and the unified bounds. Note that when all coordinates have the same smoothness, that is \( \alpha_{1} = \cdots = \alpha_{d} = \alpha \), we have \( \alphabar = \alpha / d \). An example with an unbounded input space and a density ratio that is unbounded in both directions is
\[
  q(x) =
  \frac{\ee^{2}}{\sqrt{\pi}}
  \exp(-\ee^{x} - \ee^{-x} - x / 2),
  \quad
  p(x) = q(-x),
  \quad
  x \in \bbR,
\]
for which \( \rstar(x) = \ee^{x} \) and \( \fstar(x) = x \). In this case, \ASMref{asm: (sec-3) density-ratio-tails} holds with \( \kratio = 2 \), \ASMref{asm: (sec-3) input-tails} with \( \kinput = 2 \), and \ASMref{asm: (sec-3) local-smoothness} with \( \alpha_{1} = 1 \), \( H_{0} = 2 \), and \( \zeta = 1 \).

We next decompose the excess risk \( \calE \) into the errors analyzed in the subsequent subsections. For any comparison function \( f \in \ClassClip \), we have
\begin{equation}
  \label{eq: (sec-3) error-decomposition}
  \begin{aligned}
    \ExRisk{\fhat_{K, B_{N}}}
     & =
    \Risk{\fhat_{K, B_{N}}} - \Risk{\fstar}
    =
    \Dparen*{
      \Risk{\fhat_{K, B_{N}}} - \Risk{f}
    } +
    \ExRisk{f}
    \\
     & =
    \Dparen*{
      \Risk{\fhat_{K, B_{N}}} -
      \RiskN{\fhat_{K, B_{N}}}
    } +
    \Dparen*{\RiskN{f} - \Risk{f}}
    \\
     & \quad +
    \ExRisk{f} +
    \Dparen*{
      \RiskN{\fhat_{K, B_{N}}} - \RiskN{f}
    }.
  \end{aligned}
\end{equation}
In \eqref{eq: (sec-3) error-decomposition}, we express \( \ExRisk{\fhat_{K, B_{N}}} \) as a sum of four terms. The first two compare the population and empirical risks of the same function. We refer to their sum as the \emph{estimation error} and control it uniformly over \( \ClassClip \). Specifically, \THMref{thm: (sec-3) estimation-error} bounds its absolute value by
\[
  \frac{1}{4} \Dparen*{
    \ExRisk{\fhat_{K, B_{N}}} + \ExRisk{f}
  }
  + \text{statistical term}.
\]
Substituting this bound into \eqref{eq: (sec-3) error-decomposition} and moving \( \ExRisk{\fhat_{K, B_{N}}} / 4 \) to the left-hand side gives
\[
  \ExRisk{\fhat_{K, B_{N}}}
  \le
  \frac{5}{3} \ExRisk{f}
  + \frac{4}{3} \family{
    \RiskN{\fhat_{K, B_{N}}} - \RiskN{f}
  }_{+}
  + \text{statistical term},
\]
where \( \family{\cdot}_{+} = \max \family{0, \cdot} \).

Here, \( \ExRisk{f} = \Risk{f} - \Risk{\fstar} \) measures how closely the comparison function \( f \) approximates \( \fstar \) in risk. We therefore call it the \emph{approximation error}. To bound this term, \THMref{thm: (sec-3) approximation-error} constructs a comparison tree \( \fapp \) with small excess risk, as detailed in \SECref{subsec: (sec-3) approximation-error}.

With \( f = \fapp \), it remains to control \( \family{\RiskN{\fhat_{K, B_{N}}} - \RiskN{\fapp}}_{+} \). Since the leaf values of \( \fapp \) are determined by the unknown \( \fstar \), we introduce another tree \( \fref \), defined in \eqref{eq: (sec-3) empirical-comparison-tree}, to remove this dependence. By construction, \( \RiskN{\fref} \le \RiskN{\fapp} \). Hence,
\[
  \family{
    \RiskN{\fhat_{K, B_{N}}} - \RiskN{\fapp}
  }_{+}
  \le
  \family{
    \RiskN{\fhat_{K, B_{N}}} - \RiskN{\fref}
  }_{+}.
\]
We call the right-hand side the \emph{optimization error} and denote it by \( \Opterror \). \THMref{thm: (sec-3) optimization-error} in \SECref{subsec: (sec-3) optimization-error} bounds this quantity.

With bounds on all three errors in place, \SECref{subsec: (sec-3) unified-bound} combines them to bound the excess risk in \THMref{thm: (sec-3) unified-finite-sample}. By choosing the tuning parameters appropriately, we further obtain an explicit convergence rate in \CORref{cor: (sec-3) explicit-rate}. In the following subsections, we present the details of this analysis.

\subsection{Estimation Error}
\label{subsec: (sec-3) estimation-error}

We now control the estimation error in \eqref{eq: (sec-3) error-decomposition}. Since both \( \fhat_{K, B_{N}} \) and the comparison function \( f \) may depend on the training samples, we need a bound that holds uniformly over \( \ClassClip \). To this end, we first express the risk through the loss at \( (X, X') \sim P \otimes Q \):
\begin{equation}
  \label{eq: (sec-3) loss-definition}
  \ell_{f}(X, X') =
  \frac{1}{2} \ee^{-f(X) / 2} +
  \frac{1}{2} \ee^{f(X') / 2}.
\end{equation}
Then \( \Risk{f} = \EE[P \otimes Q]{\ell_{f}(X, X')} \). We compare this loss with that of \( \fstar_{B_{N}} = \TN \fstar \), which minimizes \( \Risk{f} \) over functions taking values in \( [-B_{N}, B_{N}] \).

We also use covering numbers to obtain the required uniform bound. For a function class \( \calA \) and a probability measure \( S \), let \( \calN(\eta, \calA, \scrL^{1}(S)) \) denote the minimum number of radius-\( \eta \) balls covering \( \calA \) in the \( \scrL^{1}(S) \) norm. Define \( \calG(B_{N}) = \family{\ee^{(f - B_{N}) / 2} \mid f \in \ClassClip} \). As \APPref{subsec: (app-a) estimation-proof} shows, in the present setting,
\begin{equation}
  \label{eq: (sec-3) function-covering-number}
  \log \calN(
  \eta, \calG(B_{N}), \scrL^{1}(S)
  ) \le
  C K L \log(d K L) (
  \log(1 / \eta) + 1
  ),
\end{equation}
for every \( L \ge 2 \), every probability measure \( S \) on \( \bbR^{d} \), and every \( 0 < \eta \le 1 \).

Using \eqref{eq: (sec-3) function-covering-number} alone would give the slower square-root rate \citep[see, e.g.,][Theorem~17.1]{Anthony1999NeuralNL}. Instead, we combine it with the following property of the balanced Hellinger risk: for every \( f \) with \( \abs{f} \le B_{N} \),
\begin{equation}
  \label{eq: (sec-3) bernstein-condition}
  \EE{
    \Dparen*{
      \ell_{f}(X, X') -
      \ell_{\fstar_{B_{N}}}(X, X')
    }^{2}
  } \le
  2 \cosh(B_{N} / 2) (
  \Risk{f} - \Risk{\fstar_{B_{N}}}
  ).
\end{equation}
Together, \eqref{eq: (sec-3) function-covering-number} and \eqref{eq: (sec-3) bernstein-condition} yield the faster rate stated below.
\begin{theorem}[Estimation error]
  \label{thm: (sec-3) estimation-error}
  Suppose that \( L \ge 2 \). For every \( 0 < \delta < 1 \), there is a universal constant \( C \) such that, with probability at least \( 1 - \delta \), the following inequality holds simultaneously for all \( \fhat, f \in \ClassClip \):
  \begin{equation}
    \label{eq: (sec-3) estimation-error-bound}
    \begin{aligned}
       & \eqspace
      \abs{
        \Dparen*{
          \Risk{\fhat} - \RiskN{\fhat}
        } +
        \Dparen*{\RiskN{f} - \Risk{f}}
      }
      \\ & \le
      \frac{1}{4} \Dparen*{
        \ExRisk{\fhat} + \ExRisk{f}
      } +
      C \ee^{B_{N} / 2}
      \frac{
        K L \log(d K L) \log N +
        \log(2 / \delta)
      }{N}.
    \end{aligned}
  \end{equation}
  Consequently, if \( \fhat, f \in \ClassClip \) and there is an \( \varepsilon > 0 \) such that
  \[
    \RiskN{\fhat} \le
    \RiskN{f} + \varepsilon,
  \]
  then
  \begin{equation}
    \label{eq: (sec-3) estimation-comparison-bound}
    \ExRisk{\fhat} \le
    \frac{5}{3} \ExRisk{f} +
    \frac{4}{3} \varepsilon +
    C \ee^{B_{N} / 2}
    \frac{
      K L \log(d K L) \log N +
      \log(2 / \delta)
    }{N}.
  \end{equation}
\end{theorem}
The left-hand side of \eqref{eq: (sec-3) estimation-error-bound} is the absolute value of the estimation error in \eqref{eq: (sec-3) error-decomposition}. In \eqref{eq: (sec-3) estimation-comparison-bound}, \( \ExRisk{f} \) is the approximation error, and \( \varepsilon \) bounds \( \family{\RiskN{\fhat} - \RiskN{f}}_{+} \) from above. The theorem requires the functions in \( \ClassClip \) to be bounded, but it does not require \( \fstar \) itself to be bounded. \APPref{subsec: (app-a) estimation-proof} gives the proof.

\subsection{Approximation Error}
\label{subsec: (sec-3) approximation-error}

We next bound the approximation error in \eqref{eq: (sec-3) error-decomposition}. To this end we construct a comparison function \( f \) from the trees that the algorithm can represent, with \( f \) close to \( \fstar \) and hence of small excess risk. Since the smoothness condition \ASMref{asm: (sec-3) local-smoothness} is stated on cubes, we build the tree inside the cube \( (-u_{N}, u_{N}]^{d} \), where \( u_{N} > 0 \) grows with \( N \), and set it to zero outside. The cube is half open to match the convention of \SECref{subsec: (sec-2) axis-aligned-newton-trees} that the region on the left of a split point \( c \) is \( \family{x \mid x^{(j)} \le c} \).

Before stating the theorem, we quantify how finely the allowed split points \( \family{\Split}_{j = 1}^{d} \) of \SECref{subsec: (sec-2) axis-aligned-newton-trees} subdivide the interval \( [-u_{N}, u_{N}] \) in each coordinate. We include \( -u_{N} \) and \( u_{N} \) in these sets and measure the \emph{resolution} of coordinate \( j \) by
\begin{equation}
  \label{eq: (sec-3) cutpoint-resolution}
  \Resolution =
  \sup_{x^{(j)} \in [-u_{N}, u_{N}]}
  \,
  \inf_{c \in \Split \cap [-u_{N}, u_{N}]}
  \abs{x^{(j)} - c}.
\end{equation}
This is the largest distance from a point in the interval to its nearest allowed split point, so smaller \( \Resolution \) means that \( \Split \) is denser in \( [-u_{N}, u_{N}] \).

\begin{theorem}[Approximation error]
  \label{thm: (sec-3) approximation-error}
  Suppose that Assumptions~\ref{asm: (sec-3) density-ratio-tails}--\ref{asm: (sec-3) local-smoothness} hold. Let \( u_{N} > 0 \), \( B_{N} > 0 \), and \( L > 2 d \). Given any realization of \( \family{\Split}_{j = 1}^{d} \), there is a tree \( \fapp \) with at most \( L \) leaves that splits only at points of these sets. It takes values in \( [-B_{N}, B_{N}] \), is zero outside \( (-u_{N}, u_{N}]^{d} \), and satisfies
  \begin{equation}
    \label{eq: (sec-3) approximation-error}
    \begin{aligned}
      \ExRisk{\fapp} \le
      C \biggg(
       &
      \ee^{B_{N} / 2}
      H_{0}^{2} (1 + u_{N})^{2 \zeta}
      \Dparen*{
        L^{-2 \alphabar} +
        \sum_{j = 1}^{d}
        \Dparen*{
          \frac{\Resolution}{u_{N}}
        }^{2 \alpha_{j}}
      }
      \\ & +
      \exp \Dparen*{
        -\frac{\ee^{B_{N}}}{2 \kratio}
      } +
      \exp \Dparen*{-\frac{u_{N}}{\kinput}}
      \biggg).
    \end{aligned}
  \end{equation}
  Here \( \alphabar^{-1} = \sum_{j = 1}^{d} \alpha_{j}^{-1} \), and \( \Resolution \) is defined in \eqref{eq: (sec-3) cutpoint-resolution}.
  If, in addition, the allowed split points are fine enough that, for some fixed \( C_{\mathrm{grid}} > 0 \) and every \( j \in \family{1, \dots, d} \),
  \begin{equation}
    \label{eq: (sec-3) grid-condition}
    \frac{\Resolution}{u_{N}} \le
    C_{\mathrm{grid}}
    L^{-\alphabar / \alpha_{j}},
  \end{equation}
  then \eqref{eq: (sec-3) approximation-error} reduces to
  \begin{equation}
    \label{eq: (sec-3) approximation-error-fine}
    \ExRisk{\fapp} \le
    C \Dparen*{
      \ee^{B_{N} / 2}
      H_{0}^{2} (1 + u_{N})^{2 \zeta}
      L^{-2 \alphabar} +
      \exp \Dparen*{
        -\frac{\ee^{B_{N}}}{2 \kratio}
      } +
      \exp \Dparen*{-\frac{u_{N}}{\kinput}}
    }.
  \end{equation}
\end{theorem}
The proof is given in \APPref{subsec: (app-a) approximation-proof}. The first line of \eqref{eq: (sec-3) approximation-error} corresponds to the finite number of leaves and to the resolution of the split points, and the second line controls the regions where \( \abs{\fstar} > B_{N} \) and where the input lies outside the cube \( (-u_{N}, u_{N}]^{d} \). Condition \eqref{eq: (sec-3) grid-condition} quantifies how fine the allowed split points must be in each coordinate: the exponent \( \alphabar / \alpha_{j} \) increases as \( \alpha_{j} \) decreases, so the less smooth coordinates require a denser \( \Split \), while smoother coordinates tolerate coarser split points.

We sketch the construction of \( \fapp \) in three steps: (1) divide the interval in each coordinate at the allowed split points, (2) merge adjacent intervals (whose Cartesian products form the leaves inside the cube) until the tree has at most \( L \) leaves, and (3) assign to every leaf inside the cube the value of \( \fstar_{B_{N}} = \TN \fstar \) at its midpoint, and \( 0 \) to the leaves outside the cube. We call \( \fapp \) the \emph{population comparison tree}.

The tree \( \fapp \) is not computable from the training samples, since its leaf values depend on the unknown \( \fstar \). We instead use the leaf partition of \( \fapp \), denoted by \( \PartionApp \), and minimize the empirical risk \eqref{eq: (sec-2) empirical-hellinger-risk} over the trees on \( \PartionApp \) that take values in \( [-B_{N}, B_{N}] \) inside the cube \( (-u_{N}, u_{N}]^{d} \) and vanish outside. Let \( \fref \) be the minimizer. Then it is given by
\begin{equation}
  \label{eq: (sec-3) empirical-comparison-tree}
  \left. \fref \right|_{A} =
  \left\{
  \begin{array}{ll}
    \displaystyle
    \TN \log \frac{
      \PP[P_{N}]{X \in A}
    }{\PP[Q_{N}]{X' \in A}},
     &
    \PP[P_{N}]{X \in A} > 0,
    \PP[Q_{N}]{X' \in A} > 0;
    \\
    B_{N},
     &
    \PP[P_{N}]{X \in A} > 0,
    \PP[Q_{N}]{X' \in A} = 0;
    \\
    -B_{N},
     &
    \PP[P_{N}]{X \in A} = 0,
    \PP[Q_{N}]{X' \in A} > 0;
    \\
    0,
     &
    \PP[P_{N}]{X \in A} = 0,
    \PP[Q_{N}]{X' \in A} = 0,
    \\
  \end{array}
  \right.
\end{equation}
on every leaf \( A \in \PartionApp \) inside the cube, and \( 0 \) on the leaves outside the cube. Thus \( \fref \) depends on the training samples but not on the unknown \( \fstar \). We call it the \emph{empirical comparison tree}. Since \( \fref \) minimizes the empirical risk over a class containing \( \fapp \), we have \( \RiskN{\fref} \le \RiskN{\fapp} \). Hence the empirical risk of \( \fhat_{K, B_{N}} \) exceeds that of \( \fapp \) by at most
\begin{equation}
  \label{eq: (sec-3) optimization-error-definition}
  \Opterror =
  \family{
    \RiskN{\fhat_{K, B_{N}}} - \RiskN{\fref}
  }_{+},
\end{equation}
which bounds the positive part of the last term in \eqref{eq: (sec-3) error-decomposition} when \( f = \fapp \). The bound \eqref{eq: (sec-3) estimation-comparison-bound} in \THMref{thm: (sec-3) estimation-error} then applies with \( f = \fapp \) and \( \varepsilon = \Opterror \), and \THMref{thm: (sec-3) approximation-error} controls its approximation term.

\subsection{Optimization Error}
\label{subsec: (sec-3) optimization-error}

It remains to bound \( \Opterror \), the only quantity in \eqref{eq: (sec-3) estimation-comparison-bound} that has not yet been controlled. Recall that \( \fhat_{K} \) is the unclipped function produced by Algorithm~\ref{alg: (sec-2) training} after \( K \) iterations and that \( \fhat_{K, B_{N}} = \TN \fhat_{K} \). We separate \( \Opterror \) into the effect of clipping and the gap between \( \fhat_{K} \) and \( \fref \):
\begin{equation}
  \label{eq: (sec-3) optimization-decomposition}
  \Opterror \le
  \family{
    \RiskN{\fhat_{K, B_{N}}} - \RiskN{\fhat_{K}}
  }_{+} +
  \family{\RiskN{\fhat_{K}} - \RiskN{\fref}}_{+}.
\end{equation}

We bound the first term of \eqref{eq: (sec-3) optimization-decomposition} by the same argument as in the proof of \THMref{thm: (sec-3) estimation-error}, applied to a different function class. Before bounding the second term of \eqref{eq: (sec-3) optimization-decomposition}, we consider the decrease in the empirical risk at a single iteration. A special property of the balanced Hellinger risk puts it on the same scale as the second-order decrease \( \Dhat_{t} \) that the algorithm greedily maximizes: the leaf values \eqref{eq: (sec-2) leaf-value} lie in \( [-2, 2] \), so the Taylor remainder is at most a constant multiple of \( \Dhat_{t} \). The next theorem states the resulting bound.
\begin{theorem}[Empirical-risk decrease]
  \label{thm: (sec-3) empirical-risk-decrease}
  Let \( 0 < \nu \le 1 \). At every iteration \( t \), the trajectory \( \family{\fhat_{t}}_{t = 0}^{K} \) produced by Algorithm~\ref{alg: (sec-2) training} satisfies
  \begin{equation}
    \label{eq: (sec-3) empirical-risk-decrease}
    (2 \nu - \nu^{2}) \Dhat_{t} \le
    \RiskN{\fhat_{t}} - \RiskN{\fhat_{t + 1}} \le
    2 \nu \Dhat_{t}.
  \end{equation}
\end{theorem}

Thus, whenever \( \Dhat_{t} > 0 \), the algorithm strictly decreases the empirical risk. For a fixed \( \Dhat_{t} \), the lower bound in \eqref{eq: (sec-3) empirical-risk-decrease} is largest at \( \nu = 1 \), but \( \Dhat_{t} \) is computed from \( \fhat_{t} \), which also depends on \( \nu \), so \( \nu = 1 \) is not necessarily optimal. With the following assumption, the bound in \THMref{thm: (sec-3) empirical-risk-decrease} controls the remaining gap to \( \fref \).
\begin{assumption}[Progress toward the comparison tree]
  \label{asm: (sec-3) progress-toward-comparison}
  For the trajectory \( \family{\fhat_{t}}_{t = 0}^{K} \) produced by Algorithm~\ref{alg: (sec-2) training} on the given samples, there is a constant \( 0 < \omega \le 1 \), independent of \( N \), such that
  \[
    2 \Dhat_{t} \ge
    \omega \Dparen*{
      \RiskN{\fhat_{t}} - \RiskN{\fref}
    },
  \]
  for every \( t = 0, \dots, K - 1 \) with \( \RiskN{\fhat_{t}} > \RiskN{\fref} \).
\end{assumption}
The balanced Hellinger risk gives \ASMref{asm: (sec-3) progress-toward-comparison} a geometric interpretation: if every observation were allowed its own increment, one Newton step would give a vector whose squared norm (induced by the Hessian) is exactly \( \RiskN{\fhat_{t}} \), and whose projection onto the vectors that are constant within each leaf of \( \Pihat_{t} \) is exactly the value vector of \( \hhat_{t} \), which has squared norm \( 2 \Dhat_{t} \). Therefore, \ASMref{asm: (sec-3) progress-toward-comparison} reduces to asking that the projection account for a fraction \( \omega \) of what is left to improve over \( \fref \).

\APPref{subsubsec: (app-a) optimization-theorem-proofs} contains the proof of \THMref{thm: (sec-3) empirical-risk-decrease}, while \APPref{subsubsec: (app-a) assumption-four-discussion} establishes the geometric properties stated above and gives a sufficient condition for \ASMref{asm: (sec-3) progress-toward-comparison}. Combining this assumption with \THMref{thm: (sec-3) empirical-risk-decrease} gives
\[
  \family{
    \RiskN{\fhat_{t + 1}} - \RiskN{\fref}
  }_{+} \le
  \Dparen*{
    1 - \frac{2 \nu - \nu^{2}}{2} \omega
  }
  \family{
    \RiskN{\fhat_{t}} - \RiskN{\fref}
  }_{+},
\]
and iterating this inequality bounds the second term in \eqref{eq: (sec-3) optimization-decomposition}. Together with the bound on the first term, this yields the following theorem on the optimization error \( \Opterror \).
\begin{theorem}[Optimization error]
  \label{thm: (sec-3) optimization-error}
  Suppose that \ASMref{asm: (sec-3) density-ratio-tails} holds. Fix \( N \ge 1 \), \( K \ge 1 \), \( L > 2 d \), \( u_{N} > 0 \), \( B_{N} > 0 \), and \( 0 < \nu \le 1 \) before training, take \( \fref \) to be the empirical comparison tree of \SECref{subsec: (sec-3) approximation-error}, and run Algorithm~\ref{alg: (sec-2) training}. Then, for every \( 0 < \delta < 1 \),
  \begin{equation}
    \label{eq: (sec-3) optimization-error-bound}
    \begin{aligned}
      \Opterror
      \le{} &
      C \Dparen*{
        \ee^{B_{N} / 2}
        \frac{
          K L \log(d K L) \log N +
          \log(4 / \delta)
        }{N} +
        \exp \Dparen*{
          -\frac{\ee^{B_{N}}}{8 \kratio}
        }
      }
      \\ & +
      \exp \Dparen*{
        -\frac{2 \nu - \nu^{2}}{2} \omega K
      }
    \end{aligned}
  \end{equation}
  holds with probability at least \( 1 - \delta \), uniformly over all trajectories satisfying \ASMref{asm: (sec-3) progress-toward-comparison}.
\end{theorem}
The first line of \THMref{thm: (sec-3) optimization-error} comes from clipping, and the second from \ASMref{asm: (sec-3) progress-toward-comparison}. The term with \( \ee^{B_{N} / 2} \) has the same order as the statistical term of \THMref{thm: (sec-3) estimation-error}. \APPref{subsubsec: (app-a) optimization-theorem-proofs} contains the proof.

\subsection{Unified Bounds and Model Selection}
\label{subsec: (sec-3) unified-bound}

We now combine the estimation, approximation, and optimization results. \THMref{thm: (sec-3) unified-finite-sample} states the resulting bound on the excess risk \( \ExRisk{\fhat_{K, B_{N}}} \).
\begin{theorem}[Unified finite-sample bound]
  \label{thm: (sec-3) unified-finite-sample}
  Suppose that Assumptions~\ref{asm: (sec-3) density-ratio-tails}--\ref{asm: (sec-3) local-smoothness} hold. Before training, fix \( N \ge 1 \), \( K \ge 1 \), \( L > 2 d \), \( u_{N} > 0 \), \( B_{N} > 0 \), and \( 0 < \nu \le 1 \). For every \( 0 < \delta < 1 \), with probability at least \( 1 - \delta \), every trajectory satisfying \ASMref{asm: (sec-3) progress-toward-comparison} obeys
  \begin{equation}
    \label{eq: (sec-3) explicit-main-bound}
    \ExRisk{\fhat_{K, B_{N}}} \le
    C (E_{1} + E_{2} + E_{3} + E_{4}),
  \end{equation}
  where
  \[
    \begin{aligned}
      E_{1}
       & =
      \ee^{B_{N} / 2} \frac{
        K L \log(d K L) \log N +
        \log(8 / \delta)
      }{N},
      \\
      E_{2}
       & =
      \ee^{B_{N} / 2}
      H_{0}^{2} (1 + u_{N})^{2 \zeta}
      \Dparen*{
        L^{-2 \alphabar} +
        \sum_{j = 1}^{d} \Dparen*{
          \frac{\Resolution}{u_{N}}
        }^{2 \alpha_{j}}
      },
      \\
      E_{3}
       & =
      \exp \Dparen*{
        -\frac{2 \nu - \nu^{2}}{2} \omega K
      },
      \quad
      E_{4} =
      \exp \Dparen*{
        -\frac{\ee^{B_{N}}}{8 \kratio}
      } +
      \exp \Dparen*{-\frac{u_{N}}{\kinput}}.
    \end{aligned}
  \]
  Without \ASMref{asm: (sec-3) progress-toward-comparison}, the bound remains valid with \( E_{3} \) replaced by the optimization error \( \Opterror \) of \eqref{eq: (sec-3) optimization-error-definition}, which does not involve \( \fstar \).
\end{theorem}
Here \( E_{1} \) is the estimation term of \THMref{thm: (sec-3) estimation-error}, \( E_{2} \) the approximation term of \THMref{thm: (sec-3) approximation-error}, \( E_{3} \) the optimization term of \THMref{thm: (sec-3) optimization-error}, and \( E_{4} \) collects the two tail terms, which bound the contributions of the regions \( \family{\abs{\fstar} > B_{N}} \) and \( \bbR^{d} \setminus (-u_{N}, u_{N}]^{d} \). The proof is in \APPref{subsec: (app-a) unified-validation-proofs}.

With a slight abuse of notation, we allow \( K \) and \( L \) to depend on \( N \). This yields an explicit convergence rate, as stated in \CORref{cor: (sec-3) explicit-rate} below.
\begin{corollary}[Explicit convergence rate]
  \label{cor: (sec-3) explicit-rate}
  Suppose that \ASMref{asm: (sec-3) progress-toward-comparison} and the conditions of \THMref{thm: (sec-3) unified-finite-sample} hold. Choose \( B_{N} \), \( u_{N} \), \( K \), and \( L \) as functions of \( N \) as follows:
  \begin{equation}
    \begin{aligned}
      \ee^{B_{N}}
       & = 8 \kratio \log N,
       & u_{N}
       & = \kinput \log N,
      \\
      K
       & = \left\lceil
      \frac{2 \log N}{
        (2 \nu - \nu^{2}) \omega
      }
      \right\rceil,
       & L
       & = \left\lceil
      \Dparen*{
        N (\log N)^{2 \zeta - 3}
      }^{1 / (2 \alphabar + 1)}
      \right\rceil,
    \end{aligned}
  \end{equation}
  and let the allowed split points be fine enough that \eqref{eq: (sec-3) grid-condition} is satisfied. Then, for every \( 0 < \delta < 1 \), with probability at least \( 1 - \delta \),
  \begin{equation}
    \label{eq: (sec-3) convergence-rate}
    \ExRisk{\fhat_{K, B_{N}}} \le
    C \Dparen*{
      N^{-\frac{2 \alphabar}{2 \alphabar + 1}}
      (\log N)^{
          \frac{
            7 \alphabar + 2 \zeta + 1 / 2
          }{2 \alphabar + 1}
        } +
      N^{-1} (\log N)^{1 / 2} \log(8 / \delta)
    }.
  \end{equation}
  The constant \( C \) is independent of \( N \) and \( \delta \).
\end{corollary}
Up to logarithmic factors, the bound \eqref{eq: (sec-3) convergence-rate} is of order \( N^{-2 \alphabar / (2 \alphabar + 1)} \), which agrees with the minimax \( \scrL^{2} \) rate for estimating a function that is \( \alpha_{j} \)-smooth along coordinate \( j \) \citep{Hoffmann2002RandomAR}. When all coordinates have the same smoothness, that is \( \alpha_{1} = \cdots = \alpha_{d} = \alpha \), we have \( \alphabar = \alpha / d \), and the bound scales as \( N^{-2 \alpha / (2 \alpha + d)} \), the minimax \( \scrL^{2} \) rate for \( \alpha \)-H\"{o}lder-smooth estimation in \( d \) dimensions \citep{Stone1982OptimalGR}. Under the stated assumptions, this rate is attained up to logarithmic factors even when both the density ratio and its reciprocal are unbounded.

Finally, we consider the additional error from selecting the tuning parameters by validation, since the schedules in \CORref{cor: (sec-3) explicit-rate} depend on constants that are not known in practice. Suppose the algorithm is run for \( M \) candidates, giving \( \fhat^{\langle 1 \rangle}, \dots, \fhat^{\langle M \rangle} \), and that \( N_{v} \) validation observations are drawn from each distribution. Denote their empirical distributions by \( P_{v} \) and \( Q_{v} \), and define the \emph{validation risk} by
\begin{equation}
  \label{eq: (sec-3) validation-risk}
  \RiskNv{f} =
  \frac{1}{2} \EE[P_{v}]{\ee^{-f(X) / 2}} +
  \frac{1}{2} \EE[Q_{v}]{\ee^{f(X') / 2}}.
\end{equation}
We select the candidate with the smallest validation risk. The next theorem bounds the excess risk of the selected model in terms of the best candidate and \( N_{v} \).
\begin{theorem}[Validation selection]
  \label{thm: (sec-3) validation-selection}
  Suppose the training and validation samples are mutually independent, with i.i.d.\ observations within each sample, and that every candidate takes values in \( [-B_{N}, B_{N}] \). Then, conditional on the training samples, for every \( 0 < \delta < 1 \),
  \begin{equation}
    \label{eq: (sec-3) validation-bound}
    \ExRisk{\fhat^{\langle \mhat \rangle}} \le
    \frac{5}{3}
    \min_{1 \le m \le M}
    \ExRisk{\fhat^{\langle m \rangle}} +
    C \ee^{B_{N} / 2}
    \frac{
      (\log M) \log N_{v} +
      \log(2 / \delta)
    }{N_{v}},
  \end{equation}
  holds with probability at least \( 1 - \delta \), where \( \mhat \) is the index of the candidate with the smallest validation risk.
\end{theorem}
The proof is in \APPref{subsec: (app-a) unified-validation-proofs}. When \( M \) is fixed and \( N_{v} \) is of the order of \( N \), the second term of \eqref{eq: (sec-3) validation-bound} is absorbed into the estimation term \( E_{1} \) in \THMref{thm: (sec-3) unified-finite-sample}, so validation selection does not change the convergence rate.

\section{Related Work and Discussion}
\label{sec: (sec-4) related-work}

The preceding sections have presented the algorithm and its finite-sample theory. We now compare our method and analysis with existing work on direct density ratio estimation and on the theory of boosting algorithms and trees. We then discuss possible extensions of this work.

\subsection{Direct Density Ratio Estimation}
\label{subsec: (sec-4) direct-density-ratio}

Most direct density ratio estimators \citep[e.g.,][]{Sugiyama2012DensityRE} minimize a convex risk over a function class fixed in advance, and their theoretical guarantees apply to an empirical optimizer over that class. In the work of \citet{Nguyen2010EstimatingDF}, the density ratio is estimated through such a convex risk, and the metric entropy of the function class governs the convergence rate. Kernel-based methods use the same convex-risk formulation over a fixed kernel expansion, including KLIEP \citep{Sugiyama2008DirectIE}, uLSIF \citep{Kanamori2009LeastSquaresAD}, RuLSIF \citep{Yamada2013RelativeDR}, and more recent estimators \citep{Gizewski2022RegularizationUD, Liu2026UnboundedDR}. Their rates depend on the kernel and on the magnitude and smoothness of the density ratio. In contrast to these methods, the tree class considered here is nonconvex, and the analysis applies to the output of the training algorithm, not to an empirical risk minimizer.

A second family of direct estimators obtains the ratio from a binary classifier. With equal class priors, the density ratio equals the ratio of the posterior probabilities that an observation comes from \( P \) and from \( Q \) \citep{Qin1998InferencesCS, Gutmann2010NoiseContrastiveEN}. The binary loss of the classifier determines the population risk of the estimator, and the exponential loss \citep[Example~1]{Zellinger2025BinaryLD} is the case used here: its population risk is the balanced Hellinger risk in \eqref{eq: (sec-2) balanced-hellinger-risk}. Most classifier-based estimators are parametric, whereas the additive tree studied here is nonparametric, and the two settings are not directly comparable.

Several recent analyses address density ratios that need not be bounded. \citet{Gogolashvili2023ImportanceWC} and \citet{Guo2025OnlineLA} use the ratio in a downstream task without estimating it, and require only moment conditions on it rather than boundedness. Among the analyses that estimate the ratio, \citet{Xu2025EstimatingUD} use a local H\"{o}lder class with sub-exponential tails, which accommodates an unbounded input space and an unbounded ratio; their result holds in expectation. \citet{Zheng2026ErrorAD} estimate the logarithm of the ratio with deep ReLU networks under a Bregman divergence; their extension to unbounded ratios requires the ratio to have a positive infimum. \citet{Liu2026UnboundedDR} estimate the unbounded density ratio by transforming it into a bounded relative density ratio in a reproducing kernel Hilbert space and then mapping the estimate back. Our work follows this line of research with a tree estimator. The ratio and its reciprocal may both be unbounded, with sub-exponential tails in both directions and a clipping threshold that grows with the sample size. Our bound holds with high probability under a local H\"{o}lder assumption (\ASMref{asm: (sec-3) local-smoothness}) similar to the one in \citet{Xu2025EstimatingUD}.

Tree-based density ratio estimation, by contrast, has received comparatively little attention. \citet{Hushchyn2021GeneralizationCD} apply gradient-boosted trees and neural networks to change-point detection, without a statistical bound for the tree class. \citet{Awaya2026TwoSampleCA} is the closest work to ours: they model the square root of the density ratio, and their population criterion equals twice the balanced Hellinger risk \eqref{eq: (sec-2) balanced-hellinger-risk}. For a fixed leaf partition, they characterize the optimal tree added in a single boosting step. Neither study analyzes the finite training trajectory that produces the estimator considered here. Our algorithm is Newton boosting rather than gradient boosting. We exploit the following property of the balanced Hellinger risk itself: its exponential form keeps the leaf values of \eqref{eq: (sec-2) leaf-value} in \( [-2, 2] \), which controls the Taylor remainder and yields bounds on the decrease of the empirical risk in \THMref{thm: (sec-3) empirical-risk-decrease}. This bound is what makes the optimization error of the realized trajectory analyzable.

\subsection{Theory of Boosting Algorithms and Trees}
\label{subsec: (sec-4) boosting-tree-theory}

The optimization analyses of boosting concern the decrease of the empirical risk on the training sample rather than the excess risk. \citet{Sun2014ConvergenceRA} show for the logistic loss that this decrease is faster with Newton steps than with gradient steps, and \citet{Telgarsky2012PrimalDualCA} obtains a similar conclusion for a broad family of losses. In practice the loss is chosen by the user, and widely used implementations such as XGBoost \citep{Chen2016XGBoostST} and LightGBM \citep{Ke2017LightGBMHE} take Newton steps derived from a second-order Taylor polynomial of the empirical risk. The experiments in \SECref{sec: (sec-5) experiments} use LightGBM. For the balanced Hellinger risk \eqref{eq: (sec-2) balanced-hellinger-risk}, the bounds in \THMref{thm: (sec-3) empirical-risk-decrease} control the empirical risk itself, going beyond the second-order Taylor polynomial. Furthermore, the optimization error of the realized trajectory enters the unified bound of \THMref{thm: (sec-3) unified-finite-sample}, which bounds the excess risk of the fitted estimator, going beyond its empirical risk.

The statistical theory for trees mostly concerns standard regression and classification tasks. \citet{Klusowski2020SparseLC} studies greedy regression trees under the squared loss and shows that the split gain is governed by the Pearson correlation between the target variable and the best split available at that node, a data-dependent measure of how much structure the split extracts. \citet{Xu2026StatisticalOO} give oracle inequalities and minimax rates for the empirically optimal tree with at most \( L \) leaves, but not for the output of the algorithm analyzed here. Our analysis assumes, for the balanced Hellinger risk, that the second-order decrease at every iteration accounts for a fixed fraction of the current empirical-risk gap (\ASMref{asm: (sec-3) progress-toward-comparison}). This assumption is imposed on the training trajectory instead of on the two distributions. The resulting bound applies to the output of the training algorithm (\THMref{thm: (sec-3) optimization-error}) and not to an empirical-risk minimizer.

\subsection{Future Work}
\label{subsec: (sec-4) future-work}

Our results leave three questions open.
\begin{itemize}
  \item \ASMref{asm: (sec-3) progress-toward-comparison} is imposed on the training trajectory, whereas the other assumptions are imposed on the two data distributions. Whether this condition on the training trajectory holds with high probability for specific data structures or suitable choices of the tuning parameters remains unknown. A positive answer would place the rate in \CORref{cor: (sec-3) explicit-rate} under conditions on the data and the algorithm alone.

  \item Our analysis relies on both the balanced Hellinger risk and the axis-aligned tree class. In particular, the proof of \THMref{thm: (sec-3) empirical-risk-decrease} uses the boundedness of the leaf values in \eqref{eq: (sec-2) leaf-value}, ensured by the form of the risk, together with the constancy of each tree on its leaves. A natural next step is to examine other combinations of risk and function class, and to determine which parts of our argument can carry over.

  \item The validation bound covers a fixed, finite candidate set. If a comparable bound could be established for tuning parameters adjusted adaptively using the validation results, it would justify reusing the validation samples throughout the tuning process.
\end{itemize}

\section{Numerical Experiments}
\label{sec: (sec-5) experiments}

This section evaluates the finite-sample theory of \SECref{sec: (sec-3) finite-sample} on synthetic data, and the performance of Hellinger trees on public datasets. In every experiment that fits additive trees, the trees are grown with LightGBM \citep{Ke2017LightGBMHE}. The allowed split points are then the thresholds that LightGBM derives from the fitting samples, and they remain fixed throughout training.

\subsection{Synthetic Experiments}
\label{subsec: (sec-5) synthetic-experiments}

The synthetic experiments examine the estimation, approximation, and optimization errors, followed by overall convergence. We use four pairs of distributions, each specified by \( Q \) and the density ratio \( \rstar = p / q \), which together determine \( P \):
\begin{itemize}
  \item%
        \emph{Equal}. \( P = Q = \operatorname{Unif}([-1, 1]^{d}) \), so \( \rstar = 1 \).

  \item%
        \emph{Bounded}. \( Q = \operatorname{Unif}([-1, 1]) \) and \( \rstar(x) = 0.8 \ee^{0.8 x} / \sinh 0.8 \).

  \item%
        \emph{Unbounded}. \( q(x) \propto \exp(-\ee^{\abs{x}}) \) and \( \rstar(x) \propto \ee^{x} \) on \( \bbR \), where the constants normalize \( q \) to a density and make \( \EE[Q]{\rstar} = 1 \).

  \item%
        \emph{Nonadditive}. \( Q = \operatorname{Unif}([-1, 1]^{2}) \) and \( \rstar(x) \propto \exp(0.8 \abs{x^{(1)} x^{(2)}}) \).
\end{itemize}
Since \( \fstar = \log \rstar \) is known, we evaluate the excess risk \( \calE \) exactly using \eqref{eq: (sec-2) cosh-excess-risk}.

\begin{figure}[ht]
  \centering
  \includegraphics{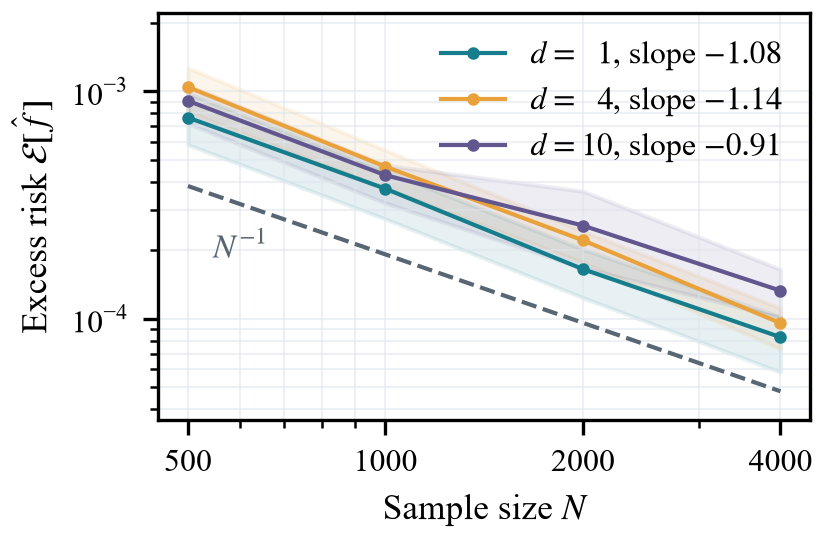}
  \caption{Estimation error with fixed numbers of trees and leaves. Each curve reports the mean and interquartile range over 20 independent fits.}
  \label{fig: (sec-5) estimation-error}
\end{figure}

\noindent \textbf{Estimation error.} The experiment uses the \emph{Equal} pair with \( d \in \family{1, 4, 10} \). Then \( \fstar = \log \rstar = 0 \), so the approximation error vanishes and the estimation error is isolated. Each sample contains \( N = 500 \), 1,000, 2,000, or 4,000 observations. Every fit runs \( K = 4 \) iterations with at most \( L = 4 \) leaves per tree and a learning rate of \( \nu = 0.1 \), and each configuration is repeated independently 20 times. With \( K \) and \( L \) fixed, \THMref{thm: (sec-3) estimation-error} gives \( N^{-1} \) as the leading power, up to logarithmic factors. The fitted slopes in \FIGref{fig: (sec-5) estimation-error} confirm this rate in all three dimensions.

\noindent \textbf{Approximation error.} The experiment uses the \emph{Bounded} pair, which is one-dimensional with a \( 1 \)-Lipschitz log density ratio, so \ASMref{asm: (sec-3) local-smoothness} holds with \( \alpha_{1} = 1 \), giving \( \alphabar = 1 \). We run two experiments, one varying the number of leaves and one varying the allowed split points.

\begin{figure}[ht]
  \centering
  \includegraphics{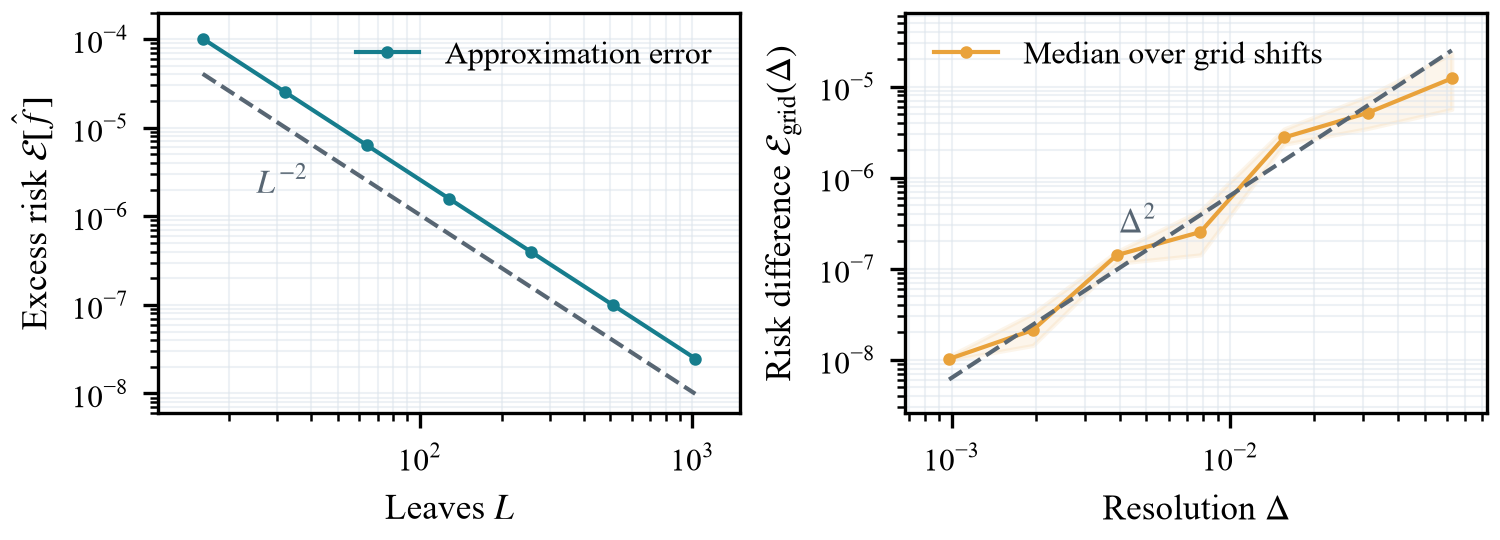}
  \caption{Approximation error. The left panel varies the number of leaves. The right panel compares two trees with \( L = 4 \) leaves, one restricted to a uniform grid of allowed split points and the other using all possible split points. Markers show the median over the 20 grids, and shading shows the interquartile range. The dashed lines have slopes \( -2 \) and \( 2 \).}
  \label{fig: (sec-5) approximation-error}
\end{figure}

In the first experiment, we divide \( [-1, 1] \) into \( L \) equal-width intervals for \( L \in \family{2^{4}, 2^{5}, \dots, 2^{10}} \), and compare \( \fstar \) with a comparison tree that is constant on each interval and takes the value of \( \fstar \) at its midpoint. Both functions are known, so the excess risk is computed exactly and no repetitions are involved. With \( \alphabar = 1 \), \THMref{thm: (sec-3) approximation-error} gives an excess risk of order \( L^{-2} \) for this comparison tree. The left panel of \FIGref{fig: (sec-5) approximation-error} confirms this rate.

In the second experiment, we take \( D \in \family{2^{4}, 2^{5}, \dots, 2^{10}} \) and place the allowed split points on a uniform grid of \( D \) intervals of \( [-1, 1] \), whose resolution is \( \Delta = 1 / D \). We keep \( -1 \) and \( 1 \) as grid points and add 19 equally spaced shifts of the interior points, which gives 20 grids for each \( D \). For every grid, we compare two quantities: the minimum risk of a tree with \( L = 4 \) leaves restricted to that grid, and the minimum risk over all possible split points. The right panel of \FIGref{fig: (sec-5) approximation-error} shows that the difference between the two risks grows like \( \Delta^{2} \), as predicted by \THMref{thm: (sec-3) approximation-error}.

\noindent \textbf{Optimization error.} We next examine how the second-order decrease \( \Dhat_{t} \) relates to the decrease in the original empirical risk \( \widehat{\calR}_{N} \). \THMref{thm: (sec-3) empirical-risk-decrease} guarantees the following bound at every iteration:
\[
  (2 \nu - \nu^{2}) \Dhat_{t} \le
  \RiskN{\fhat_{t}} - \RiskN{\fhat_{t + 1}} \le
  2 \nu \Dhat_{t}.
\]
For each of the four pairs of distributions  defined above, we use \( N \in \family{500, 1000, 2000, 4000} \), with five repetitions at each sample size. Every fit runs \( K = 50 \) iterations with at most \( L = 4 \) leaves per tree and a learning rate of \( \nu = 0.1 \), giving lower and upper coefficients of 0.19 and 0.20. At each iteration, we record \( \Dhat_{t} \) before the update and \( \RiskN{\fhat_{t}} - \RiskN{\fhat_{t + 1}} \) after the update.

\begin{figure}[ht]
  \centering
  \includegraphics{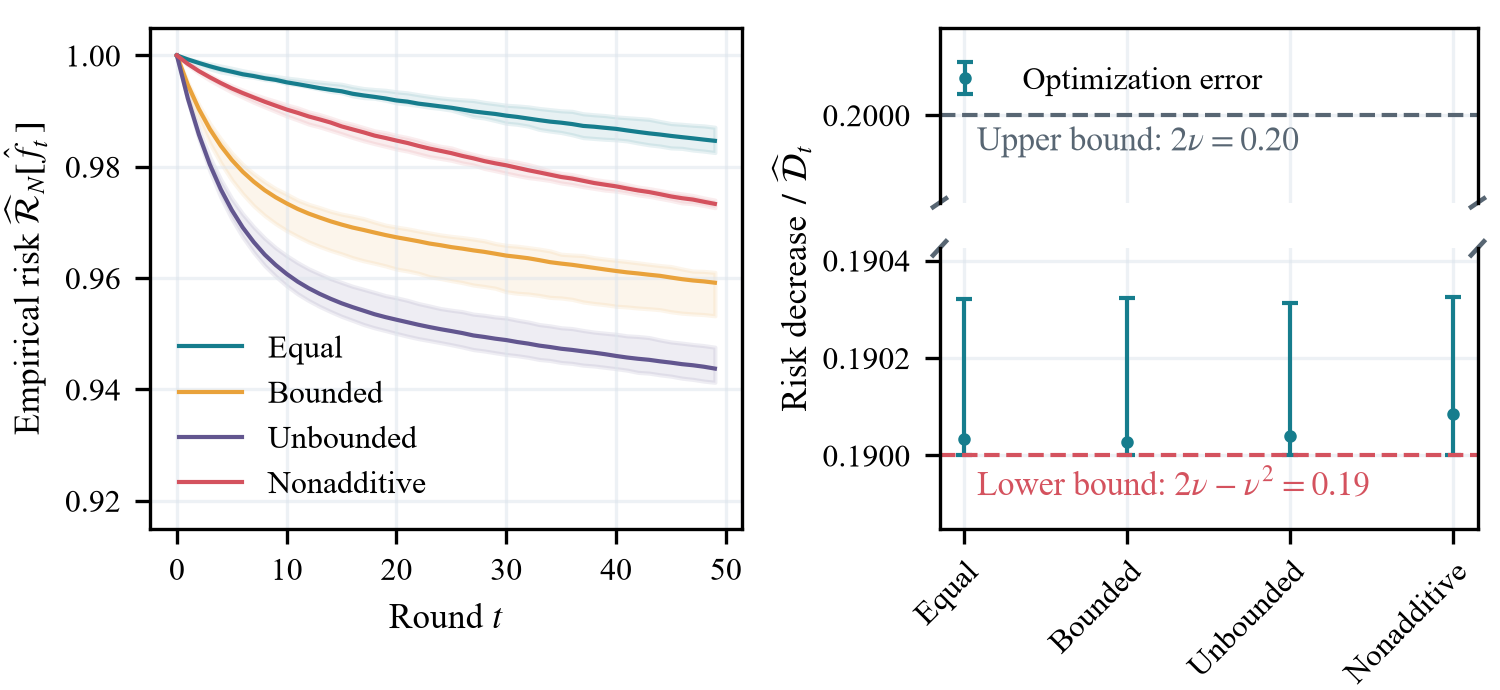}
  \caption{Decrease in the original empirical risk. Left: mean and interquartile range over five independent fits at \( N = 1000 \). Right: median and full range of the empirical-risk decrease relative to \( \Dhat_{t} \), pooled over all iterations and four sample sizes. The dashed lines show the bounds in \THMref{thm: (sec-3) empirical-risk-decrease}.}
  \label{fig: (sec-5) optimization-error}
\end{figure}

As shown in \FIGref{fig: (sec-5) optimization-error}, every computed value of \( (\RiskN{\fhat_{t}} - \RiskN{\fhat_{t + 1}}) / \Dhat_{t} \) lies between 0.19 and 0.20. Thus, the numerical results satisfy the theoretical bound at every tested iteration. These values concentrate near 0.19, showing that the lower bound is tight in all four pairs.

\noindent \textbf{Overall convergence.} We now examine overall convergence when all three errors are present. The experiment uses the \emph{Bounded} and \emph{Unbounded} pairs: both are one-dimensional, and their log density ratios are \( 1 \)-Lipschitz, so \ASMref{asm: (sec-3) local-smoothness} holds with \( \alpha_{1} = 1 \), giving \( \alphabar = 1 \). \CORref{cor: (sec-3) explicit-rate} then shows that the excess risk decreases at rate \( N^{-2 / 3} \), up to logarithmic factors. We use sample sizes \( N \in \family{4^{5}, 4^{6}, \dots, 4^{9}} \), each with 20 repetitions, and we set the learning rate to \( \nu = 0.1 \) and the number of iterations to \( K = 640 \), 768, 896, 1,024, and 1,152, which scales with \( \log N \). To check that the scale of the leaf limit does not affect the convergence rate, we consider the two choices
\[
  L = \left\lceil
  \Dparen*{\frac{N}{\log N}}^{1 / 3}
  \right\rceil
  \quad \text{or} \quad
  L = \left\lceil
  2 \Dparen*{\frac{N}{\log N}}^{1 / 3}
  \right\rceil.
\]
The clipping threshold is \( B_{N} = \log(96 \log N) \) in \emph{Bounded} and \( B_{N} = \log(160 \log N) \) in \emph{Unbounded}, in line with the tail constants of the two pairs, and both use \( u_{N} = 4 \log N \).

\begin{figure}[ht]
  \centering
  \includegraphics{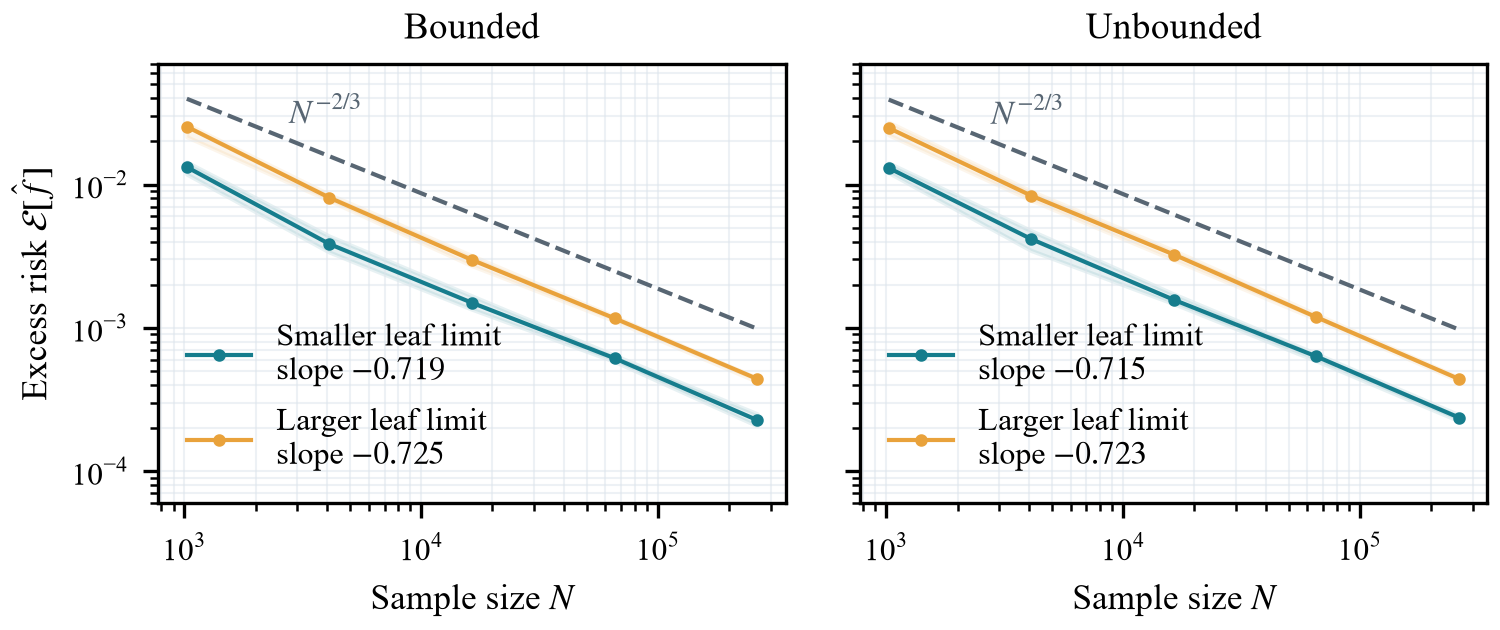}
  \caption{Excess risk when all three errors are present. Each panel shows the two choices of the leaf limit. Markers show means over 20 independent fits, and shading shows the interquartile ranges. The dashed lines have slope \( -2 / 3 \).}
  \label{fig: (sec-5) combined-errors}
\end{figure}

As shown in \FIGref{fig: (sec-5) combined-errors}, the four empirical log--log slopes across the two pairs are close to \( -2 / 3 \), in agreement with the theoretical rate.

\subsection{Public-Data Experiments}
\label{subsec: (sec-5) real-data}

These experiments assess the performance of Hellinger trees against four baseline methods on 10 tasks built from public datasets: eight from the UCI Machine Learning Repository \citep{Kelly2024UCIML} and two from OpenML \citep{Bischl2025OpenML}. Each task treats two groups of observations of one dataset as samples from \( P \) and \( Q \). Since \( \fstar \) is unknown, the excess risk cannot be computed. We instead report the empirical risk \( \widehat{\calR}_{N} \) in \eqref{eq: (sec-2) empirical-hellinger-risk} on two independent test samples, one from each distribution, which are used only for the final evaluation.

For each task, we repeat the experiment independently 20 times. In each repetition, the observations of each group are split at random into fitting, validation, and test samples in the proportions \( 60\% \), \( 20\% \), and \( 20\% \), except that Avila and COIL 2000 keep the official test set and split the official training set \( 80\% \) and \( 20\% \) between fitting and validation. The two fitting samples are reduced to a common size by random subsampling. We write \( N \) for the number of fitting observations per group, and \( P_{N} \) and \( Q_{N} \) for the empirical distributions of the two subsamples.

Preprocessing of the raw features uses only the fitting samples: missing values of numeric features are filled with the median, categorical features are one-hot encoded, and all features are standardized to zero mean and unit variance. Task details are given in \TABref{tab: (sec-5) real-data-tasks}. The five methods share the same observations for fitting, validation, and testing, and the same preprocessing.

\begin{table}[ht]
  \centering
  \small
  \caption{Public-data tasks. Here \( N \) is the number of fitting observations per group after the two fitting samples are balanced, and \( d \) is the input dimension after encoding, reported as a range when it varies with the categories present in the fitting sample.}
  \label{tab: (sec-5) real-data-tasks}
  \begin{tabular}{lllrrr}
    \toprule
    Task
     & \( P \)
     & \( Q \)
     & Raw features
     & \( d \)
     & \( N \)
    \\
    \midrule
    \multicolumn{6}{l}{\emph{UCI Machine Learning Repository} \citep{Kelly2024UCIML}}
    \\[0.5em]
    Avila
     & copyist A
     & copyist F
     & 10
     & 10
     & 1,568
    \\
    COIL 2000
     & class 1
     & class 0
     & 85
     & 85
     & 278
    \\
    Credit Default
     & default
     & no default
     & 23
     & 23
     & 3,981
    \\
    Electrical Grid
     & stable
     & unstable
     & 12
     & 12
     & 2,172
    \\
    Letter
     & A--M
     & N--Z
     & 16
     & 16
     & 5,964
    \\
    Madelon
     & class \( -1 \)
     & class \( +1 \)
     & 500
     & 500
     & 600
    \\
    Online Shoppers
     & revenue
     & no revenue
     & 17
     & 69--74
     & 1,144
    \\
    Student Dropout
     & dropout
     & graduate
     & 36
     & 202--215
     & 852
    \\[0.5em]
    \multicolumn{6}{l}{\emph{OpenML} \citep{Bischl2025OpenML}}
    \\[0.5em]
    Airline Satisfaction
     & satisfied
     & dissatisfied
     & 21
     & 86--89
     & 35,275
    \\
    Telco Churn
     & churn
     & no churn
     & 19
     & 46
     & 1,121
    \\
    \bottomrule
  \end{tabular}
\end{table}

\noindent \textbf{Candidate methods.} We compare Hellinger trees with four baselines, two that estimate the density ratio directly (uLSIF and KLIEP) and two that estimate it from a binary classifier (logistic trees and kernel logistic regression). Under equal class priors, the latter two recover the density ratio as the posterior probability of the \( P \) label divided by that of the \( Q \) label \citep{Qin1998InferencesCS, Menon2016LinkingLD}. The methods and their candidate parameters are as follows:
\begin{itemize}
  \item%
        \emph{Hellinger trees} (ours). This method follows Algorithm~\ref{alg: (sec-2) training}, which uses Newton boosting for the balanced Hellinger risk. The candidates are all combinations of a leaf limit \( L \in \family{4, 8, 16, 32} \) and a pair \( (\nu, K) \) with \( \nu \in \family{0.01, 0.1, 1} \) and \( K \in \family{2, 4, 8, 16} / \nu \). Varying the learning rate \( \nu \) and the number of iterations \( K \) jointly is standard practice in boosting \citep{Friedman2001GreedyFA}.

  \item%
        \emph{Logistic trees} \citep{Friedman2001GreedyFA}. This method uses the same settings as Hellinger trees, but replaces the empirical Hellinger risk \( \widehat{\calR}_{N} \) in \eqref{eq: (sec-2) empirical-hellinger-risk} by the empirical logistic risk:
        \begin{equation}
          \label{eq: (sec-5) logistic-risk}
          \widehat{\calL}_{N}[f] =
          \frac{1}{2} \EE[P_{N}]{
            \log (1 + \ee^{-f(X)})
          } +
          \frac{1}{2} \EE[Q_{N}]{
            \log (1 + \ee^{f(X')})
          }.
        \end{equation}

  \item%
        \emph{Kernel logistic regression} \citep{Zhu2005KernelLR}. This and the following two kernel methods use the same Gaussian kernel \( k_{\sigma}(x, c) = \exp(-\norm{x - c}_{2}^{2} / (2 \sigma^{2})) \). They also use the same centers \( c_{1}, \dots, c_{M_{\mathrm{ker}}} \) (with \( M_{\mathrm{ker}} \le 50 \)), drawn without replacement from the two fitting samples. Kernel logistic regression minimizes the empirical logistic risk \eqref{eq: (sec-5) logistic-risk} over
        \[
          f_{\mathrm{KLR}}(x) =
          \theta_{0} +
          \theta_{\mathrm{lin}}^{\top} x +
          \theta_{\mathrm{ker}}^{\top} \bsk_{\sigma}(x),
          \quad
          \bsk_{\sigma}(z) = (k_{\sigma}(z, c_{1}), \dots, k_{\sigma}(z, c_{M_{\mathrm{ker}}}))^{\top},
        \]
        with the regularizer \( \lambda ( \norm{\theta_{\mathrm{lin}}}_{2}^{2} + \norm{\theta_{\mathrm{ker}}}_{2}^{2} ) / 2 \) on the linear and kernel coefficients. The candidates are all pairs \( (\sigma, \lambda) \): the bandwidth \( \sigma \) is one of the 8 equally spaced quantiles from \( 0.05 \) to \( 0.95 \) of the pairwise distances within the fitting samples, based on the median heuristic for the kernel width \citep{Gretton2012KernelTT}, and the regularization value \( \lambda \) spans the wide range \( \family{10^{-4}, 10^{-3}, \dots, 10} \).

  \item%
        \emph{uLSIF} \citep{Kanamori2009LeastSquaresAD}. This method estimates the density ratio through the kernel expansion
        \[
          r_{\theta}(x) =
          \sum_{l = 1}^{M_{\mathrm{ker}}} \theta_{l} \, k_{\sigma}(x, c_{l})
        \]
        and minimizes the regularized least-squares objective
        \[
          \min_{\theta} \family{
            \frac{1}{2} \theta^{\top} \widehat{H} \theta -
            \widehat{h}^{\top} \theta +
            \frac{\lambda}{2} \norm{\theta}_{2}^{2}
          },
        \]
        where \( c_{l} \), \( \sigma \), and \( \lambda \) are the same as in kernel logistic regression. The matrix and vector of the objective are
        \[
          \widehat{H} =
          \frac{1}{N}
          \sum_{i = 1}^{N}
          \bsk_{\sigma}(X'_{i})
          \bsk_{\sigma}(X'_{i})^{\top},
          \quad
          \widehat{h} =
          \frac{1}{N}
          \sum_{i = 1}^{N}
          \bsk_{\sigma}(X_{i}),
        \]
        and we solve for \( \theta \), setting negative coefficients to zero to obtain a nonnegative estimator.

  \item%
        \emph{KLIEP} \citep{Sugiyama2008DirectIE}. This method uses the same kernel expansion as uLSIF, but a different objective:
        \[
          \max_{\theta \ge 0}
          \frac{1}{N} \sum_{i = 1}^{N}
          \log \Dparen*{
            \theta^{\top} \bsk_{\sigma}(X_{i})
          },
          \quad
          \text{subject to }
          \frac{1}{N}
          \sum_{i = 1}^{N}
          \theta^{\top} \bsk_{\sigma}(X'_{i}) = 1.
        \]
        The candidates of KLIEP are the 48 bandwidths \( \sigma \) given by the equally spaced quantiles from \( 0.05 \) to \( 0.95 \) of the pairwise distances within the fitting samples.
\end{itemize}
Every method has 48 candidates.

\noindent \textbf{Model selection and evaluation.} All candidates are fitted on the two fitting samples, and every estimated log density ratio is clipped to \( [-8, 8] \) before the validation and test risks are computed. For uLSIF and KLIEP, the clipping applies to the logarithm of the estimate. We use the same selection rule for the five methods, retaining the candidate with the smallest validation risk \eqref{eq: (sec-3) validation-risk} and evaluating the retained model on the two test samples.

\begin{figure}[ht]
  \centering
  \includegraphics{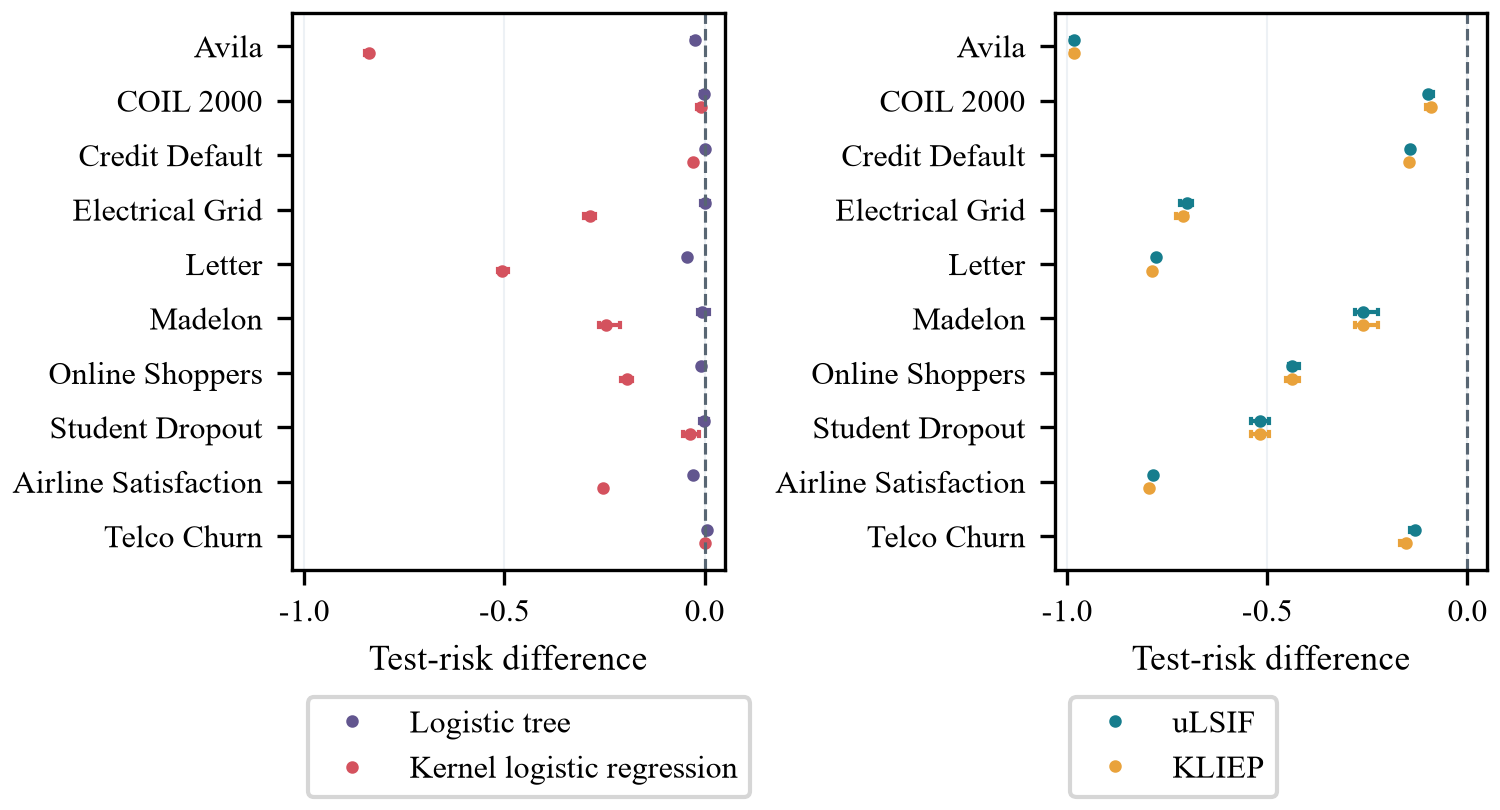}
  \caption{Comparison of Hellinger trees with four baseline methods on the 10 public-data tasks. For each task, a marker shows the mean difference in test risk between Hellinger trees and a baseline method over 20 independent repetitions, and the error bar shows the interquartile range. Negative values favor Hellinger trees.}
  \label{fig: (sec-5) real-data}
\end{figure}

The results in \TABref{tab: (sec-5) real-data-results} show that Hellinger trees have the lowest average test risk among the five methods. Their performance is close to that of logistic trees, with a smaller average test risk on 7 of the 10 tasks. They outperform kernel logistic regression on 9 tasks, and both uLSIF and KLIEP on all 10 tasks. \FIGref{fig: (sec-5) real-data} shows, for each task, the difference between the test risk of Hellinger trees and that of each baseline method.

\begin{table}[ht]
  \centering
  \caption{Average test Hellinger risk over the 10 tasks. Parentheses contain the standard deviation over the individual runs, and the last column gives the number of tasks on which Hellinger trees have the smaller average risk.}
  \label{tab: (sec-5) real-data-results}
  \begin{tabular}{lcc}
    \toprule
    Method
     & Test Hellinger risk
     & Tasks favoring Hellinger trees
    \\
    \midrule
    \textbf{Hellinger trees (ours)}
     & \( \mathbf{0.5108 \, (\pm 0.3003)} \)
     & --
    \\
    Logistic trees
     & \( 0.5210 \, (\pm 0.2891) \)
     & \( 7 / 10 \)
    \\
    Kernel logistic regression
     & \( 0.7503 \, (\pm 0.1732) \)
     & \( 9 / 10 \)
    \\
    uLSIF
     & \( 0.9942 \, (\pm 0.0321) \)
     & \( 10 / 10 \)
    \\
    KLIEP
     & \( 0.9994 \, (\pm 0.0221) \)
     & \( 10 / 10 \)
    \\
    \bottomrule
  \end{tabular}
\end{table}

\appendix
\section{Proofs}
\label{sec: (app-a) proofs}

This appendix contains the proofs of the results in \SECref{sec: (sec-3) finite-sample}. Throughout, \( C \) may vary between occurrences, but it does not depend on \( N \) or \( \delta \).

\subsection{Proof of the Estimation Error}
\label{subsec: (app-a) estimation-proof}

This subsection proves \THMref{thm: (sec-3) estimation-error}. To that end, we first state the following lemma, which bounds the deviation between the empirical and the population expectations uniformly over a class of bounded functions, with a remainder that depends on its covering numbers. It is also used in \APPref{subsec: (app-a) optimization-proof} and \ref{subsec: (app-a) unified-validation-proofs}.

\begin{lemma}
  \label{lem: (app-a) peeled-relative-deviation}
  Let \( W \) be a probability distribution, and write \( W_{N} \) for its empirical distribution over i.i.d.\ observations \( Z_{1}, \dots, Z_{N} \). Let \( \calA \) be a class of functions with \( 0 \in \calA \) such that every \( g \in \calA \) satisfies \( \abs{g} \le 1 \) and \( \EE[W]{g^{2}} \le \EE[W]{g} \). Suppose that, for some \( V \ge 1 \), the covering numbers satisfy
  \begin{equation}
    \label{eq: (app-a) covering-number}
    \log \calN(\eta, \calA, \scrL^{1}(S)) \le
    V (\log(1 / \eta) + 1)
  \end{equation}
  for every probability measure \( S \) and every \( 0 < \eta \le 1 \). Then, for every \( 0 < \delta < 1 \), the following inequality holds with probability at least \( 1 - \delta \):
  \begin{equation}
    \label{eq: (app-a) peeled-relative-deviation}
    \abs{\EE[W_{N}]{g} - \EE[W]{g}}
    \le
    \frac{1}{4} \EE[W]{g} +
    C \frac{
      V \log N + \log(2 / \delta)
    }{N},
    \quad
    \forall g \in \calA,
  \end{equation}
  where \( C \) is a universal constant.
\end{lemma}

\begin{proof}
  We prove the lemma by applying \citet[Theorem~3.3, part~1]{Bartlett2005LocalRC}. That theorem requires a bounded class whose variance is controlled by its mean, and a sub-root function \( \psi \) that dominates, for every \( s > 0 \), the expected Rademacher averages of the subclasses \( \calA(s) = \family{g \in \calA \mid \EE[W]{g^{2}} \le s} \). The hypotheses of the lemma give the first requirement; the second is met below.

  Conditional on the observations, write \( \widehat{\frakR}_{N}(s) \) for the empirical Rademacher average of \( \calA(s) \) and \( \hat{\rho}_{N}(s) \) for the largest \( \scrL^{2}(W_{N}) \) norm of the same class:
  \[
    \widehat{\frakR}_{N}(s) =
    \EE[\sigma]{
      \sup_{g \in \calA(s)}
      \abs{
        \frac{1}{N}
        \sum_{i = 1}^{N} \sigma_{i} g(Z_{i})
      }
    },
    \quad
    \hat{\rho}_{N}(s) =
    \Dparen*{
      \sup_{g \in \calA(s)} \EE[W_{N}]{g^{2}}
    }^{1/2},
  \]
  where the independent signs \( \sigma_{i} \) take values \( \pm 1 \) with equal probability, and \( 0 \in \calA \) keeps every subclass \( \calA(s) \) nonempty.

  We bound \( \EE{\widehat{\frakR}_{N}(s)} \) through covering numbers. By \eqref{eq: (app-a) covering-number}, there is a net \( \calH(s) \subseteq \calA(s) \) of radius \( 1 / N \) in the \( \scrL^{1}(W_{N}) \) norm such that \( \log \abs{\calH(s)} \le C V \log N \). Hence
  \begin{equation}
    \label{eq: (app-a) rademacher-net}
    \begin{aligned}
      \widehat{\frakR}_{N}(s)
       & =
      \EE[\sigma]{
        \sup_{g \in \calA(s)}
        \abs{
          \frac{1}{N}
          \sum_{i = 1}^{N} \sigma_{i} g(Z_{i})
        }
      } \overset{\text{(a)}}{\le}
      \EE[\sigma]{
        \sup_{h \in \calH(s)}
        \abs{
          \frac{1}{N}
          \sum_{i = 1}^{N} \sigma_{i} h(Z_{i})
        }
      } +
      \frac{1}{N}
      \\ &
      \overset{\text{(b)}}{\le}
      4 \hat{\rho}_{N}(s)
      \sqrt{\frac{\log \abs{\calH(s)}}{N}} +
      \frac{1}{N} \le
      C \hat{\rho}_{N}(s)
      \sqrt{\frac{V \log N}{N}} +
      \frac{1}{N},
    \end{aligned}
  \end{equation}
  where (a) replaces each \( g \in \calA(s) \) with its nearest net point \( h \in \calH(s) \), so the resulting change in the average is at most \( 1 / N \); and (b) is the Rademacher bound of a finite class of functions \citep[Lemma~4.14]{Wainwright2019HighDimensionalSN}. Furthermore,
  \begin{equation}
    \label{eq: (app-a) radius-bound}
    \begin{aligned}
      \EE{\hat{\rho}_{N}(s)^{2}}
       & =
      \EE{
        \sup_{g \in \calA(s)}
        \EE[W_{N}]{g^{2}}
      }
      \le
      s + \EE{
        \sup_{g \in \calA(s)}
        \abs{\EE[W_{N}]{g^{2}} - \EE[W]{g^{2}}}
      }
      \\
       & \le
      s + 8 \EE{\widehat{\frakR}_{N}(s)},
    \end{aligned}
  \end{equation}
  where the last inequality follows from symmetrization and the contraction property for the square function \citep[Lemma~A.5 and Theorem~A.6]{Bartlett2005LocalRC}. Taking expectations in \eqref{eq: (app-a) rademacher-net} and combining the result with \eqref{eq: (app-a) radius-bound}, we solve for \( \EE{\widehat{\frakR}_{N}(s)} \):
  \begin{equation}
    \label{eq: (app-a) rademacher-expectation}
    \EE{\widehat{\frakR}_{N}(s)} \le
    \underbrace{
      C \Dparen*{\sqrt{\frac{s V \log N}{N}} +
        \frac{V \log N}{N}}
    }_{\psi(s)}.
  \end{equation}
  We write \( \psi(s) \) for the right-hand side of \eqref{eq: (app-a) rademacher-expectation}. A direct computation shows that \( \psi \) is sub-root in the sense of \citet[Definition~3.1]{Bartlett2005LocalRC}, with positive fixed point at most \( C V \log N / N \).

  Now we apply \citet[Theorem~3.3, part~1]{Bartlett2005LocalRC} with its class and distribution replaced by our \( \calA \) and \( W \), with \( K = 6 \) and \( x = \log(2 / \delta) \), and with the fixed point of \( \psi \) bounded by \( C V \log N / N \). Its two conclusions, combined by a union bound, give
  \[
    \EE[W]{g} \le
    \frac{6}{5} \EE[W_{N}]{g} +
    C \frac{V \log N + \log(2 / \delta)}{N},
    \quad
    \EE[W_{N}]{g} \le
    \frac{7}{6} \EE[W]{g} +
    C \frac{V \log N + \log(2 / \delta)}{N},
  \]
  with probability at least \( 1 - \delta \). This yields \eqref{eq: (app-a) peeled-relative-deviation}, completing the proof.
\end{proof}

We will apply \LEMref{lem: (app-a) peeled-relative-deviation} to prove \THMref{thm: (sec-3) estimation-error}. Before doing so, we prove the covering number bound \eqref{eq: (sec-3) function-covering-number}.

\begin{proofof}{\eqref{eq: (sec-3) function-covering-number}}
  We first bound the pseudodimension of \( \ClassClip \), then use it to bound the covering numbers. To begin, the proof of \citet[Theorem~3.14]{Gupta2016PACAA} shows that, for \( L \ge 2 \), a single tree with at most \( L \) leaves can assign \( n \) inputs among them in at most \( (C d n)^{L} \) different ways, hence at most \( (C d n)^{K L} \) ways for \( K \) such trees.

  Now fix inputs \( x_{1}, \dots, x_{n} \in \bbR^{d} \), thresholds \( r_{1}, \dots, r_{n} \in \bbR \), and one of these ways. Let \( \theta \in \bbR^{K L} \) collect the \( K L \) leaf values, and write \( f_{\theta} \in \TreeClass \) for the sum of the \( K \) trees. Once we fix which leaf contains each \( x_{i} \) in each tree, \( f_{\theta}(x_{i}) \) is an affine function of \( \theta \). Thus, for each \( i \), the hyperplane \( \family{\theta \mid f_{\theta}(x_{i}) = r_{i}} \) splits \( \bbR^{K L} \) into two half-spaces. For \( n \ge K L \), the \( n \) such hyperplanes divide \( \bbR^{K L} \) into at most \( (C n / (K L))^{K L} \) disjoint regions \citep[Theorem~8.3]{Anthony1999NeuralNL}. Within each region, changing \( \theta \) does not change \( \operatorname{sgn}(f_{\theta}(x_{i}) - r_{i}) \) for \( i = 1, \dots, n \).

  Combining the preceding bounds, we obtain that for fixed inputs and thresholds, the number of distinct sign vectors \( (\operatorname{sgn}(f_{\theta}(x_{1}) - r_{1}), \dots, \operatorname{sgn}(f_{\theta}(x_{n}) - r_{n})) \) is at most \( (C d n)^{K L} (C n / (K L))^{K L} \). By definition, for \( n = \Pdim(\TreeClass) \), some choice of inputs and thresholds realizes all \( 2^{n} \) sign vectors. Consequently,
  \[
    2^{n} \le
    (C d n)^{K L}
    \Dparen*{\frac{C n}{K L}}^{K L}.
  \]
  Solving the inequality gives \( n = \Pdim(\TreeClass) \le C K L \log(d K L) \). Clipping does not increase the pseudodimension, so \( \Pdim(\ClassClip) \le \Pdim(\TreeClass) \). The strictly increasing transformation \( v \mapsto \ee^{(v - B_{N}) / 2} \) preserves the pseudodimension, so \( \Pdim(\calG(B_{N})) = \Pdim(\ClassClip) \). Every function in \( \calG(B_{N}) \) takes values in \( [0, 1] \). Thus, by \citet[Corollary~3]{Haussler1995SpherePN},
  \[
    \begin{aligned}
       & \eqspace
      \log \calN(
      \eta, \calG(B_{N}), \scrL^{1}(S)
      )
      \le
      C \Pdim(\calG(B_{N})) \,
      (\log(1 / \eta) + 1)
      \\ & \le
      C \Pdim(\TreeClass) \,
      (\log(1 / \eta) + 1)
      \le
      C K L \log(d K L)
      (\log(1 / \eta) + 1).
    \end{aligned}
  \]
  This proves \eqref{eq: (sec-3) function-covering-number}, completing the proof.
\end{proofof}

Next, we prove \eqref{eq: (sec-3) bernstein-condition}, which provides the moment condition required by \LEMref{lem: (app-a) peeled-relative-deviation}.

\begin{proofof}{\eqref{eq: (sec-3) bernstein-condition}}
  We set
  \begin{equation}
    \label{eq: (app-a) psi}
    \Psi_{f}(x) =
    \frac{1}{2} p(x)
    \Dparen*{
      \ee^{-f(x) / 2} -
      \ee^{-\fstar_{B_{N}}(x) / 2}
    }^{2} +
    \frac{1}{2} q(x)
    \Dparen*{
      \ee^{f(x) / 2} -
      \ee^{\fstar_{B_{N}}(x) / 2}
    }^{2}.
  \end{equation}
  By the definition of \( \ell_{f} \) in \eqref{eq: (sec-3) loss-definition}, we have
  \[
    \EE[P \otimes Q]{
      \Dparen*{
        \ell_{f}(X, X') -
        \ell_{\fstar_{B_{N}}}(X, X')
      }^{2}
    } \le
    \int \Psi_{f}(x) \, \dd \mu(x).
  \]
  Next, recall from \eqref{eq: (sec-2) pointwise-risk} that \( \Risk{f} = \int \Phi_{f}(x) \, \dd \mu(x) \). Hence
  \[
    \Risk{f} - \Risk{\fstar_{B_{N}}} =
    \int \Dparen*{
      \Phi_{f}(x) - \Phi_{\fstar_{B_{N}}}(x)
    } \dd \mu(x).
  \]
  Therefore, to prove \eqref{eq: (sec-3) bernstein-condition}, it suffices to prove
  \begin{equation}
    \label{eq: (app-a) pointwise-bernstein}
    \Psi_{f}(x) \le
    2 \cosh(B_{N} / 2)
    \Dparen*{
      \Phi_{f}(x) - \Phi_{\fstar_{B_{N}}}(x)
    },
    \quad
    \forall x \in \bbR^{d},
  \end{equation}
  and integrate both sides. Indeed, by an elementary calculation,
  \[
    \Psi_{f}(x) \le
    \Dparen*{
      \ee^{f(x) / 2} + \ee^{-f(x) / 2}
    }
    \Dparen*{
      \Phi_{f}(x) - \Phi_{\fstar_{B_{N}}}(x)
    }.
  \]
  Since \( \abs{f(x)} \le B_{N} \), the factor \( \ee^{f(x) / 2} + \ee^{-f(x) / 2} \) is at most \( 2 \cosh(B_{N} / 2) \). Hence \eqref{eq: (app-a) pointwise-bernstein} holds. This completes the proof.
\end{proofof}

Having established both ingredients, we now prove \THMref{thm: (sec-3) estimation-error}.

\begin{proofof}{\THMref{thm: (sec-3) estimation-error}}
  We set \( Z_{i} = (X_{i}, X'_{i}) \), with distribution \( W = P \otimes Q \). Define
  \begin{equation}
    \label{eq: (app-a) normalized-loss-difference}
    g_{f} = \frac{
      \ell_{f} - \ell_{\fstar_{B_{N}}}
    }{2 \ee^{B_{N} / 2}}
  \end{equation}
  for \( f \in \ClassClip \), and let \( \calA = \family{g_{f} \mid f \in \ClassClip} \cup \family{0} \). Each \( g_{f}(x, x') \) is obtained by adding two functions from \( \calG(B_{N}) \) evaluated at \( x \) and \( x' \), subtracting a fixed function, and scaling. These transformations do not increase the order of the logarithmic covering numbers \citep[Theorem~10]{Attias2024FatShatteringDA}. Thus, \eqref{eq: (sec-3) function-covering-number} gives
  \[
    \log \calN(\eta, \calA, \scrL^{1}(S)) \le
    C K L \log(d K L)
    (\log(1 / \eta) + 1)
  \]
  for every probability measure \( S \) on \( \bbR^{d} \times \bbR^{d} \) and every \( 0 < \eta \le 1 \). Furthermore, the functions \( g_{f} \) satisfy
  \[
    \abs{g_{f}} \le 1,
    \quad
    \EE[W]{g_{f}^{2}} \le \EE[W]{g_{f}},
  \]
  where the second inequality follows from \eqref{eq: (sec-3) bernstein-condition} and \( 2 \cosh(B_{N} / 2) \le 2 \ee^{B_{N} / 2} \). Applying \LEMref{lem: (app-a) peeled-relative-deviation} to \( g_{f} \in \calA \) and multiplying by \( 2 \ee^{B_{N} / 2} \), we obtain
  \[
    \begin{aligned}
       & \eqspace
      \abs{
        \Dparen*{
          \Risk{f} - \Risk{\fstar_{B_{N}}}
        } -
        \Dparen*{
          \RiskN{f} - \RiskN{\fstar_{B_{N}}}
        }
      }
      \\
       & \le
      \frac{1}{4} \Dparen*{
        \Risk{f} - \Risk{\fstar_{B_{N}}}
      } +
      C \ee^{B_{N} / 2}
      \frac{
        K L \log(d K L) \log N +
        \log(2 / \delta)
      }{N},
    \end{aligned}
  \]
  with probability at least \( 1 - \delta \), for every \( f \in \ClassClip \). We evaluate this inequality at \( \fhat \) and \( f \) and combine the two bounds using the triangle inequality. The terms involving \( \fstar_{B_{N}} \) cancel. Together with \( \Risk{h} - \Risk{\fstar_{B_{N}}} \le \ExRisk{h} \) for both \( h = \fhat \) and \( h = f \), we obtain \eqref{eq: (sec-3) estimation-error-bound}, which proves the first assertion of \THMref{thm: (sec-3) estimation-error}. The second assertion, \eqref{eq: (sec-3) estimation-comparison-bound}, follows by expanding the absolute value in \eqref{eq: (sec-3) estimation-error-bound}, using the assumed bound \( \RiskN{\fhat} \le \RiskN{f} + \varepsilon \) to eliminate \( \RiskN{\fhat} - \RiskN{f} \), and rearranging.
\end{proofof}

\subsection{Proof of the Approximation Error}
\label{subsec: (app-a) approximation-proof}

We prove \THMref{thm: (sec-3) approximation-error} by constructing \( \fapp \) and bounding its excess risk on three regions.

\begin{proofof}{\THMref{thm: (sec-3) approximation-error}}
  Using the definition of \( \Phi_{f} \) in \eqref{eq: (sec-2) pointwise-risk}, we write
  \begin{equation}
    \ExRisk{\fapp} =
    \int_{\bbR^{d}} \Dparen*{
      \Phi_{\fapp} - \Phi_{\fstar}
    } \dd \mu \le
    \sum_{i = 1}^{3}
    \int_{\Omega_{i}}
    \Dparen*{
      \Phi_{\fapp} - \Phi_{\fstar}
    } \dd \mu.
  \end{equation}
  The inequality follows from \( \Phi_{\fapp}(x) - \Phi_{\fstar}(x) \ge 0 \). Here,
  \[
    \Omega_{1} =
    (-u_{N}, u_{N}]^{d} \cap
    \family{\abs{\fstar} \le B_{N}},
    \quad
    \Omega_{2} =
    \family{\abs{\fstar} > B_{N}},
    \quad
    \Omega_{3} =
    \bbR^{d} \setminus
    (-u_{N}, u_{N}]^{d}.
  \]
  It suffices to bound the three integrals on the right-hand side.

  \begin{itemize}
    \item%
          To bound the integral over \( \Omega_{1} \), we construct the leaf partition \( \PartionApp \) and specify the value of \( \fapp \) on each leaf. For each \( j \in \family{1, \dots, d} \), set \( J_{j} = \lfloor (L - 2 d)^{\alphabar / \alpha_{j}} \rfloor \), which gives
          \begin{equation}
            \label{eq: (app-a) leaf-partition-count}
            \prod_{j = 1}^{d} J_{j} \le
            L - 2 d,
            \quad
            \sum_{j = 1}^{d}
            J_{j}^{-\alpha_{j}} \le
            C L^{-\alphabar}.
          \end{equation}
          In coordinate \( j \), \eqref{eq: (sec-3) cutpoint-resolution} implies that each interval between consecutive allowed split points in \( [-u_{N}, u_{N}] \) has length at most \( 2 \Resolution \). We merge these intervals from left to right, starting a new one whenever adding the next would make the current one longer than \( 2 u_{N} / J_{j} + 2 \Resolution \). This produces at most \( J_{j} \) intervals, each of length at most \( 2 u_{N} / J_{j} + 2 \Resolution \). The Cartesian products of the merged intervals give at most \( \prod_{j = 1}^{d} J_{j} \le L - 2 d \) regions inside the cube \( (-u_{N}, u_{N}]^{d} \). The region outside the cube contributes at most \( 2 d \) leaves. Together, these regions form the leaf partition \( \PartionApp \) of a tree with at most \( L \) leaves, and every split uses an allowed point.

          We set \( \fapp = 0 \) on the leaves outside \( (-u_{N}, u_{N}]^{d} \). For each leaf \( A \in \PartionApp \) contained in \( (-u_{N}, u_{N}]^{d} \), let \( x_{A} \) be its midpoint and assign the value \( \fstar_{B_{N}}(x_{A}) \) to that leaf. Every \( x \in A \) then satisfies \( \abs{x^{(j)} - x_{A}^{(j)}} \le u_{N} / J_{j} + \Resolution \) for all \( j \in \family{1, \dots, d} \). Since clipping is \( 1 \)-Lipschitz, \ASMref{asm: (sec-3) local-smoothness} gives
          \begin{equation}
            \label{eq: (app-a) leaf-approximation}
            \begin{aligned}
              \abs{
                \fapp(x) - \fstar_{B_{N}}(x)
              }
               & \le
              \abs{
                \fstar(x_{A}) - \fstar(x)
              }
              \le
              H_{0} (1 + u_{N})^{\zeta}
              \sum_{j = 1}^{d}
              \Dparen*{
                \frac{1}{2 J_{j}} +
                \frac{\Resolution}{2 u_{N}}
              }^{\alpha_{j}}
              \\
               & \le
              C H_{0} (1 + u_{N})^{\zeta}
              \Dparen*{
                L^{-\alphabar} +
                \sum_{j = 1}^{d}
                \Dparen*{
                  \frac{\Resolution}{u_{N}}
                }^{\alpha_{j}}
              }.
            \end{aligned}
          \end{equation}
          The last inequality uses \( (a + b)^{\alpha_{j}} \le a^{\alpha_{j}} + b^{\alpha_{j}} \) for \( 0 < \alpha_{j} \le 1 \), followed by \eqref{eq: (app-a) leaf-partition-count}.

          For \( x \in \Omega_{1} \), both \( \fapp(x) \) and \( \fstar(x) \) lie in \( [-B_{N}, B_{N}] \). We can therefore apply Taylor's formula to \( \Phi_{f}(x) \) at its minimizer \( \fstar(x) = \fstar_{B_{N}}(x) \), giving
          \[
            \begin{aligned}
               & \eqspace
              \Phi_{\fapp}(x) - \Phi_{\fstar}(x)
              \\ & =
              \Phi_{\fapp}(x) - \Phi_{\fstar_{B_{N}}}(x)
              \le
              \Dparen*{
                \sup_{\abs{f(x)} \le B_{N}}
                \frac{1}{2}
                \frac{
                  \partial^{2} \Phi_{f}(x)
                }{\partial f(x)^{2}}
              }
              \abs{
                \fapp(x) - \fstar_{B_{N}}(x)
              }^{2}
              \\ & \le
              \frac{
              \ee^{B_{N} / 2} (p(x) + q(x))
              }{16}
              \abs{
                \fapp(x) - \fstar_{B_{N}}(x)
              }^{2}.
            \end{aligned}
          \]
          Integrating over \( \Omega_{1} \) and using \eqref{eq: (app-a) leaf-approximation}, we obtain
          \begin{equation}
            \label{eq: (app-a) omega-one-bound}
            \int_{\Omega_{1}}
            \Dparen*{
              \Phi_{\fapp} - \Phi_{\fstar}
            } \dd \mu \le
            C \ee^{B_{N} / 2}
            H_{0}^{2} (1 + u_{N})^{2 \zeta}
            \Dparen*{
              L^{-2 \alphabar} +
              \sum_{j = 1}^{d}
              \Dparen*{\frac{\Resolution}{u_{N}}}^{2 \alpha_{j}}
            }.
          \end{equation}

    \item%
          On \( \Omega_{2} \), the bounds \( \abs{\fapp} \le B_{N} \) and \( \Phi_{\fstar} \ge 0 \) give
          \begin{equation}
            \label{eq: (app-a) omega-two-risk}
            \begin{aligned}
              \int_{\Omega_{2}}
              \Dparen*{
                \Phi_{\fapp} - \Phi_{\fstar}
              } \dd \mu
               & \le
              \int_{\Omega_{2}}
              \Phi_{\fapp} \, \dd \mu
              \le
              \frac{\ee^{B_{N} / 2}}{2}
              \int_{\Omega_{2}}
              (p + q) \, \dd \mu
              \\ & =
              \frac{\ee^{B_{N} / 2}}{2}
              (
              \PP[P]{\Omega_{2}} +
              \PP[Q]{\Omega_{2}}
              ).
            \end{aligned}
          \end{equation}
          We decompose \( \PP[P]{\Omega_{2}} \) as \( \PP[P]{(\rstar)^{-1} > \ee^{B_{N}}} + \PP[P]{\rstar > \ee^{B_{N}}} \). For the first term, Markov's inequality and \ASMref{asm: (sec-3) density-ratio-tails} give
          \[
            \PP[P]{
              (\rstar)^{-1} > \ee^{B_{N}}
            } \le
            2 \exp \Dparen*{
              -\frac{\ee^{B_{N}}}{\kratio}
            }.
          \]
          For the second term, we obtain
          \begin{equation}
            \label{eq: (app-a) density-ratio-upper-tail}
            \begin{aligned}
               & \eqspace
              \PP[P]{\rstar > \ee^{B_{N}}} =
              \EE[P]{
                \ind{
                  \family{\rstar > \ee^{B_{N}}}
                }
              }
              \overset{\text{(a)}}{=}
              \EE[Q]{
                \rstar \ind{
                  \family{\rstar > \ee^{B_{N}}}
                }
              }
              \\
               & \overset{\text{(b)}}{\le}
              (\ee^{B_{N}} + \kratio)
              \EE[Q]{
                \exp \Dparen*{
                  \frac{
                    \rstar - \ee^{B_{N}}
                  }{\kratio}
                }
              }
              \overset{\text{(c)}}{\le}
              2 (\ee^{B_{N}} + \kratio)
              \exp \Dparen*{
                -\frac{\ee^{B_{N}}}{\kratio}
              }.
            \end{aligned}
          \end{equation}
          Here, (a) follows from the definition of the density ratio; (b) uses the elementary inequality \( x \ind{\family{x > a}} \le (a + b) \exp((x - a) / b) \), valid for \( x \in \bbR \) and \( a, b > 0 \), with \( x = \rstar \), \( a = \ee^{B_{N}} \), and \( b = \kratio \); and (c) follows again from \ASMref{asm: (sec-3) density-ratio-tails}. Combining the two bounds yields
          \[
            \PP[P]{\Omega_{2}} \le
            2 (\ee^{B_{N}} + \kratio + 1)
            \exp \Dparen*{
              -\frac{\ee^{B_{N}}}{\kratio}
            }.
          \]
          The same bound holds for \( \PP[Q]{\Omega_{2}} \), so \eqref{eq: (app-a) omega-two-risk} implies
          \begin{equation}
            \label{eq: (app-a) omega-two-bound}
            \begin{aligned}
              \int_{\Omega_{2}}
              \Dparen*{\Phi_{\fapp} - \Phi_{\fstar}} \, \dd \mu
               & \le
              2 \ee^{B_{N} / 2}
              (\ee^{B_{N}} + \kratio + 1)
              \exp \Dparen*{
                -\frac{\ee^{B_{N}}}{\kratio}
              }
              \\ & \le
              C \exp \Dparen*{
                -\frac{\ee^{B_{N}}}{2 \kratio}
              }.
            \end{aligned}
          \end{equation}

    \item%
          On \( \Omega_{3} \), \( \fapp = 0 \). Hence
          \begin{equation}
            \label{eq: (app-a) omega-three-bound}
            \begin{aligned}
               & \eqspace
              \int_{\Omega_{3}}
              \Dparen*{
                \Phi_{\fapp} - \Phi_{\fstar}
              } \dd \mu
              \le
              \int_{\Omega_{3}}
              \Phi_{\fapp} \, \dd \mu
              =
              \frac{1}{2}
              \int_{\Omega_{3}}
              (p + q) \, \dd \mu
              \\ & \le
              \frac{1}{2}
              \Dparen*{
                \PP[P]{
                  \norm{X}_{\infty} \ge u_{N}
                } +
                \PP[Q]{
                  \norm{X'}_{\infty} \ge u_{N}
                }
              }
              \le
              2 \exp \Dparen*{
                -\frac{u_{N}}{\kinput}
              },
            \end{aligned}
          \end{equation}
          where the last inequality follows from Markov's inequality and \ASMref{asm: (sec-3) input-tails}.
  \end{itemize}
  Combining \eqref{eq: (app-a) omega-one-bound}, \eqref{eq: (app-a) omega-two-bound}, and \eqref{eq: (app-a) omega-three-bound} proves \eqref{eq: (sec-3) approximation-error} in \THMref{thm: (sec-3) approximation-error}. If condition \eqref{eq: (sec-3) grid-condition} holds, \eqref{eq: (sec-3) approximation-error-fine} follows directly. This completes the proof.
\end{proofof}

\subsection{Proof of the Optimization Error}
\label{subsec: (app-a) optimization-proof}

We prove the two theorems in \SECref{subsec: (sec-3) optimization-error} and then discuss \ASMref{asm: (sec-3) progress-toward-comparison}.

\subsubsection{Proofs of Theorems}
\label{subsubsec: (app-a) optimization-theorem-proofs}

To prove \THMref{thm: (sec-3) empirical-risk-decrease}, we compare the decrease in the empirical risk at each iteration with \( \Dhat_{t} \).

\begin{proofof}{\THMref{thm: (sec-3) empirical-risk-decrease}}
  Recall from \eqref{eq: (sec-2) leaf-risk} that \( \phihat_{t, A}(z) \) is the contribution of leaf \( A \) to the empirical risk when the increment on that leaf is \( z \). Since \( \fhat_{t + 1} = \fhat_{t} + \nu \hhat_{t} \) and \( \hhat_{t} \) takes the value \( \thetahat_{t}(A) \) on each leaf \( A \in \Pihat_{t} \),
  \[
    \RiskN{\fhat_{t}} -
    \RiskN{\fhat_{t + 1}} =
    \sum_{A \in \Pihat_{t}}
    \Dparen*{
      \phihat_{t, A}(0) -
      \phihat_{t, A}(\nu \thetahat_{t}(A))
    }.
  \]
  By \eqref{eq: (sec-2) leaf-derivatives} and \eqref{eq: (sec-2) second-order-decrease}, the second-order decrease \( \Dhat_{t} \) for the selected leaf partition \( \Pihat_{t} \) is
  \[
    \Dhat_{t} =
    \sum_{A \in \Pihat_{t}}
    \frac{\phihat'_{t, A}(0)^{2}}{
      2 \phihat''_{t, A}(0)
    },
  \]
  where the summand is taken to be zero when \( \phihat''_{t, A}(0) = 0 \). Thus, to establish \eqref{eq: (sec-3) empirical-risk-decrease} in \THMref{thm: (sec-3) empirical-risk-decrease}, it suffices to show that each \( A \in \Pihat_{t} \) satisfies
  \begin{equation}
    \label{eq: (app-a) leafwise-risk-decrease}
    (2 \nu - \nu^{2})
    \frac{\phihat'_{t, A}(0)^{2}}{
      2 \phihat''_{t, A}(0)
    }
    \le
    \phihat_{t, A}(0) -
    \phihat_{t, A}(\nu \thetahat_{t}(A))
    \le
    2 \nu
    \frac{\phihat'_{t, A}(0)^{2}}{
      2 \phihat''_{t, A}(0)
    }.
  \end{equation}
  Expanding \( \phihat_{t, A} \) and using the definition of \( \thetahat_{t}(A) \) in \eqref{eq: (sec-2) leaf-value} shows that \eqref{eq: (app-a) leafwise-risk-decrease} follows from
  \[
    \frac{2 \nu - \nu^{2}}{2}
    \Dparen*{\frac{\thetahat_{t}(A)}{2}}^{2}
    \le
    1 - \cosh \Dparen*{
      \nu
      \frac{\thetahat_{t}(A)}{2}
    } +
    \frac{\thetahat_{t}(A)}{2}
    \sinh \Dparen*{
      \nu
      \frac{\thetahat_{t}(A)}{2}
    }
    \le
    \nu
    \Dparen*{
      \frac{\thetahat_{t}(A)}{2}
    }^{2}.
  \]
  Since \( \abs{\thetahat_{t}(A)} \le 2 \) and \( 0 < \nu \le 1 \), this inequality always holds. Hence \eqref{eq: (app-a) leafwise-risk-decrease} holds, completing the proof of \THMref{thm: (sec-3) empirical-risk-decrease}.
\end{proofof}

To prove \THMref{thm: (sec-3) optimization-error}, we bound the two terms in the decomposition of \( \Opterror \) in \eqref{eq: (sec-3) optimization-decomposition} separately, using \ASMref{asm: (sec-3) progress-toward-comparison} for the second.

\begin{proofof}{\THMref{thm: (sec-3) optimization-error}}
  We first control the clipping term \( \family{\RiskN{\fhat_{K, B_{N}}} - \RiskN{\fhat_{K}}}_{+} \) in \eqref{eq: (sec-3) optimization-decomposition}. For any \( f \in \TreeClass \), clipping affects the empirical risk only through observations in \( \family{f > B_{N}} \cup \family{f < -B_{N}} \). More precisely, for every \( x \in \bbR^{d} \), we have
  \[
    \begin{aligned}
      \ee^{-\TN f(x) / 2} - \ee^{-f(x) / 2}
       & =
      \frac{1}{2} \int_{B_{N}}^{\infty}
      \Dparen*{
        \ee^{-u / 2} \ind{\family{f(x) > u}} -
        \ee^{u / 2} \ind{\family{f(x) < -u}}
      } \dd u,
      \\
      \ee^{\TN f(x) / 2} - \ee^{f(x) / 2}
       & =
      \frac{1}{2} \int_{B_{N}}^{\infty}
      \Dparen*{
        \ee^{-u / 2} \ind{\family{f(x) < -u}} -
        \ee^{u / 2} \ind{\family{f(x) > u}}
      } \dd u.
    \end{aligned}
  \]
  Substituting these identities into the definition of the empirical risk in \eqref{eq: (sec-2) empirical-hellinger-risk} yields
  \begin{equation}
    \label{eq: (app-a) layer-cake-clipping}
    \begin{aligned}
       & \eqspace
      \RiskN{\TN f} - \RiskN{f}
      \\ & =
      \frac{1}{2} \EE[P_{N}]{
        \ee^{-\TN f / 2} - \ee^{-f / 2}
      } +
      \frac{1}{2} \EE[Q_{N}]{
        \ee^{\TN f / 2} - \ee^{f / 2}
      }
      \\ & =
      \frac{1}{4}
      \int_{B_{N}}^{\infty}
      \ee^{-u / 2} \Bigl(
      \EE[P_{N}]{\ind{\family{f > u}}} -
      \ee^{u} \EE[Q_{N}]{\ind{\family{f > u}}}
      \\ &
      \hspace{7em} +
      \EE[Q_{N}]{\ind{\family{f < -u}}} -
      \ee^{u} \EE[P_{N}]{\ind{\family{f < -u}}}
      \Bigr) \, \dd u.
    \end{aligned}
  \end{equation}
  For \( u \ge B_{N} \), we apply \LEMref{lem: (app-a) peeled-relative-deviation} to bound \( \EE[P_{N}]{\ind{\family{f > u}}} - \ee^{u} \EE[Q_{N}]{\ind{\family{f > u}}} \). Consider the class
  \[
    \calA = \family{
      x \mapsto \ind{\family{f(x) > u}}
      \MID
      f \in \TreeClass, u \in \bbR
    } =
    \family{
      x \mapsto \ind{\family{f(x) > 0}}
      \MID
      f \in \TreeClass
    }.
  \]
  The second equality follows because \( f-u \in \TreeClass \) for every \( f \in \TreeClass \) and \( u \in \bbR \). Every \( g \in \calA \) is an indicator, so \( \abs{g} \le 1 \) and \( \EE[W]{g^{2}} = \EE[W]{g} \) for \( W \in \family{P, Q} \). This class contains the zero function, and its pseudodimension is bounded by that of \( \TreeClass \). We can therefore follow the argument in the proof of \eqref{eq: (sec-3) function-covering-number} in \APPref{subsec: (app-a) estimation-proof} to obtain
  \[
    \log \calN(\eta, \calA, \scrL^{1}(S)) \le
    C K L \log(d K L) (\log(1 / \eta) + 1)
  \]
  for every probability measure \( S \) on \( \bbR^{d} \) and every \( 0 < \eta \le 1 \). Thus the conditions of \LEMref{lem: (app-a) peeled-relative-deviation} hold.

  Applying the lemma under \( P \) and \( Q \), with failure probability \( \delta / 2 \) for each, we obtain, with probability at least \( 1 - \delta \),
  \begin{equation}
    \label{eq: (app-a) vc-relative-bound}
    \EE[W_{N}]{g} \le
    2 \EE[W]{g} + C \Gamma(N, \delta),
    \quad
    \EE[W]{g} \le
    2 \EE[W_{N}]{g} + C \Gamma(N, \delta),
  \end{equation}
  where
  \[
    \Gamma(N, \delta) =
    \frac{
      K L \log(d K L) \log N +
      \log(4 / \delta)
    }{N}.
  \]
  These bounds hold simultaneously for all \( g \in \calA \) and \( W \in \family{P, Q} \). Fix \( f \in \TreeClass \) and \( u \ge B_{N} \), and set \( g = \ind{\family{f > u}} \). On this event,
  \begin{equation}
    \label{eq: (app-a) empirical-change-of-measure}
    \begin{aligned}
      \EE[P_{N}]{g} - \ee^{u} \EE[Q_{N}]{g}
       & \overset{\text{(a)}}{\le}
      2 \EE[P]{g} -
      \ee^{u} \EE[Q_{N}]{g} +
      C \Gamma(N, \delta)
      \\ & \overset{\text{(b)}}{\le}
      \frac{\ee^{B_{N}}}{2}
      \EE[Q]{g} -
      \ee^{u} \EE[Q_{N}]{g} +
      C \Gamma(N, \delta) +
      2 \EE[Q]{
        \rstar
        \ind{\family{\rstar > \ee^{B_{N}} / 4}}
      }
      \\ & \overset{\text{(c)}}{\le}
      \underbrace{
        (\ee^{B_{N}} - \ee^{u})
        \EE[Q_{N}]{g}
      }_{\text{nonpositive}} +
      C \ee^{B_{N}} \Gamma(N, \delta) +
      2 \EE[Q]{
        \rstar
        \ind{\family{\rstar > \ee^{B_{N}} / 4}}
      }
      \\ & \overset{\text{(d)}}{\le}
      C \ee^{B_{N}} \Gamma(N, \delta) +
      (\ee^{B_{N}} + 4 \kratio)
      \exp \Dparen*{
        -\frac{\ee^{B_{N}}}{4 \kratio}
      }.
    \end{aligned}
  \end{equation}
  In (a), we apply the first bound in \eqref{eq: (app-a) vc-relative-bound} with \( W = P \). For (b), the definition of the density ratio and \( 0 \le g \le 1 \) give
  \[
    \EE[P]{g} =
    \EE[Q]{\rstar g} \le
    \frac{\ee^{B_{N}}}{4} \EE[Q]{g} +
    \EE[Q]{
      \rstar
      \ind{\family{\rstar > \ee^{B_{N}} / 4}}
    }.
  \]
  In (c), we apply the second bound in \eqref{eq: (app-a) vc-relative-bound} with \( W = Q \). In (d), the first term is nonpositive, and we bound the remaining \( \bbE_{Q} \) term as in \eqref{eq: (app-a) density-ratio-upper-tail}.

  By interchanging \( P \) and \( Q \), the same bound holds for \( \EE[Q_{N}]{\ind{\family{f < -u}}} - \ee^{u} \EE[P_{N}]{\ind{\family{f < -u}}} \). Substituting both bounds into
  \eqref{eq: (app-a) layer-cake-clipping} gives
  \begin{equation}
    \label{eq: (app-a) clipping-uniform-bound}
    \begin{aligned}
      \RiskN{\TN f} - \RiskN{f}
       & \le
      \frac{1}{2}
      \Dparen*{
        C \ee^{B_{N}}
        \Gamma(N, \delta) +
        (\ee^{B_{N}} + 4 \kratio)
        \exp \Dparen*{
          -\frac{\ee^{B_{N}}}{4 \kratio}
        }
      }
      \int_{B_{N}}^{\infty} \ee^{-u / 2} \, \dd u
      \\
       & \le
      C \Dparen*{
        \ee^{B_{N} / 2}
        \Gamma(N, \delta) +
        \exp \Dparen*{
          -\frac{\ee^{B_{N}}}{8 \kratio}
        }
      }.
    \end{aligned}
  \end{equation}
  Setting \( f = \fhat_{K} \) and taking the positive part bounds the first term in \eqref{eq: (sec-3) optimization-decomposition}.

  We next control \( \family{\RiskN{\fhat_{K}} - \RiskN{\fref}}_{+} \), the second term in \eqref{eq: (sec-3) optimization-decomposition}. When \( \RiskN{\fhat_{t}} > \RiskN{\fref} \),
  \THMref{thm: (sec-3) empirical-risk-decrease} and
  \ASMref{asm: (sec-3) progress-toward-comparison} give
  \[
    \begin{aligned}
      \family{\RiskN{\fhat_{t + 1}} -
        \RiskN{\fref}}_{+}
       & \le
      \family{
        \RiskN{\fhat_{t}} -
        \RiskN{\fref} -
        (2 \nu - \nu^{2}) \Dhat_{t}
      }_{+}
      \\ & \le
      \Dparen*{
        1 - \frac{2 \nu - \nu^{2}}{2} \omega
      }
      \family{
        \RiskN{\fhat_{t}} - \RiskN{\fref}
      }_{+}.
    \end{aligned}
  \]
  If \( \RiskN{\fhat_{t}} \le \RiskN{\fref} \), then \( \RiskN{\fhat_{t + 1}} \le \RiskN{\fref} \), since the empirical risk is nonincreasing by \THMref{thm: (sec-3) empirical-risk-decrease}. The displayed inequalities still hold, with equality at zero. Iterating the resulting bound from \( \RiskN{\fhat_{0}} = 1 \) gives
  \begin{equation}
    \label{eq: (app-a) unclipped-optimization-bound}
    \family{
      \RiskN{\fhat_{K}} - \RiskN{\fref}
    }_{+} \le
    \Dparen*{
      1 - \frac{2 \nu - \nu^{2}}{2} \omega
    }^{K} \le
    \exp \Dparen*{
      -\frac{2 \nu - \nu^{2}}{2}
      \omega K
    }.
  \end{equation}
  This bounds the second term in \eqref{eq: (sec-3) optimization-decomposition}.

  Applying \eqref{eq: (app-a) clipping-uniform-bound} and \eqref{eq: (app-a) unclipped-optimization-bound} to the two terms of \eqref{eq: (sec-3) optimization-decomposition} completes the proof of \THMref{thm: (sec-3) optimization-error}.
\end{proofof}

\subsubsection{Discussion of \ASMref{asm: (sec-3) progress-toward-comparison}}
\label{subsubsec: (app-a) assumption-four-discussion}

We establish the geometric properties of the empirical risk that motivate \ASMref{asm: (sec-3) progress-toward-comparison}, then give a sufficient condition and an example.

\begin{proofof}{the geometric properties}
  We show that the unrestricted Newton step, obtained by allowing a separate increment at each of the \( 2 N \) training observations, has squared norm \( \RiskN{\fhat_{t}} \) with respect to the inner product induced by the Hessian. We also show that, when this step is projected onto the subspace of vectors that are constant on each leaf of \( \Pihat_{t} \), the resulting vector has squared norm \( 2 \Dhat_{t} \) with respect to the same inner product.

  Define the value vector of \( f \) on the \( 2 N \) training observations by
  \[
    \operatorname{vec}(f) =
    \Dparen*{
      f(X_{1}), \dots, f(X_{N}),
      f(X'_{1}), \dots, f(X'_{N})
    }^{\top}.
  \]
  We now view \( \RiskN{f} \) in \eqref{eq: (sec-2) empirical-hellinger-risk} as a function of \( \operatorname{vec}(f) \). At \( \fhat_{t} \), its gradient \( \widehat{\sfG}_{t} \in \bbR^{2 N} \) and Hessian \( \widehat{\sfH}_{t} \in \bbR^{2 N \times 2 N} \) are given by
  \[
    \begin{aligned}
      \widehat{\sfG}_{t}
       & =
      \left.
      \nabla_{\operatorname{vec}(f)} \RiskN{f}
      \right|_{f = \fhat_{t}}
      \\
       & = \frac{1}{4 N}
      \Dparen*{
        -\ee^{-\fhat_{t}(X_{1}) / 2}, \dots,
        -\ee^{-\fhat_{t}(X_{N}) / 2},
        \ee^{\fhat_{t}(X'_{1}) / 2}, \dots,
        \ee^{\fhat_{t}(X'_{N}) / 2}
      }^{\top},
      \\
      \widehat{\sfH}_{t}
       & =
      \left.
      \nabla^{2}_{\operatorname{vec}(f)} \RiskN{f}
      \right|_{f = \fhat_{t}}
      \\
       & = \frac{1}{8 N}
      \operatorname{diag}\Dparen*{
        \ee^{-\fhat_{t}(X_{1}) / 2}, \dots,
        \ee^{-\fhat_{t}(X_{N}) / 2},
        \ee^{\fhat_{t}(X'_{1}) / 2}, \dots,
        \ee^{\fhat_{t}(X'_{N}) / 2}
      }.
    \end{aligned}
  \]
  The unrestricted Newton step is therefore
  \[
    \bsv^{\star} =
    -\widehat{\sfH}_{t}^{-1} \,
    \widehat{\sfG}_{t} =
    \big(\,
    \underbrace{2, \dots, 2}_{N}, \,
    \underbrace{-2, \dots, -2}_{N} \,
    \big)^{\top}.
  \]
  This vector is independent of \( t \). It cannot be realized by any function \( f \) if \( X_{i} = X'_{j} \) for some \( i \) and \( j \), because its coordinates prescribe different increments at the same input.

  Since \( \widehat{\sfH}_{t} \) is positive definite, it induces the following inner product and norm on \( \bbR^{2 N} \):
  \[
    \inner{\bsv_{1}, \bsv_{2}}_{\fhat_{t}} =
    \bsv_{1}^{\top} \,
    \widehat{\sfH}_{t} \,
    \bsv_{2},
    \quad
    \norm{\bsv}_{\fhat_{t}}^{2} =
    \inner{\bsv, \bsv}_{\fhat_{t}}.
  \]
  In particular,
  \begin{equation}
    \norm{\bsv^{\star}}_{\fhat_{t}}^{2} =
    (\bsv^{\star})^{\top} \,
    \widehat{\sfH}_{t} \,
    \bsv^{\star} =
    \frac{1}{2 N}
    \sum_{i = 1}^{N}
    \Dparen*{
      \ee^{-\fhat_{t}(X_{i}) / 2} +
      \ee^{\fhat_{t}(X'_{i}) / 2}
    } =
    \RiskN{\fhat_{t}}.
  \end{equation}
  This proves the first claim.

  For the second claim, fix a tree with leaf partition \( \Pi \), and write \( h = \sum_{A \in \Pi} \theta(A) \ind{A} \). Then
  \begin{equation}
    \label{eq: (app-a) newton-square-completion}
    \begin{aligned}
       & \eqspace
      \sum_{A \in \Pi}
      \Dparen*{
        \Ghat_{t}(A) \theta(A) +
        \frac{1}{2} \Hhat_{t}(A) \theta(A)^{2}
      } =
      \widehat{\sfG}_{t}^{\top} \,\operatorname{vec}(h) +
      \frac{1}{2}
      \operatorname{vec}(h)^{\top} \,
      \widehat{\sfH}_{t} \,
      \operatorname{vec}(h)
      \\ & =
      -(\bsv^{\star})^{\top} \,
      \widehat{\sfH}_{t} \,
      \operatorname{vec}(h) +
      \frac{1}{2}
      \operatorname{vec}(h)^{\top} \,
      \widehat{\sfH}_{t} \,
      \operatorname{vec}(h) =
      \frac{1}{2}
      \norm{\operatorname{vec}(h) - \bsv^{\star}}_{\fhat_{t}}^{2} -
      \frac{1}{2}
      \norm{\bsv^{\star}}_{\fhat_{t}}^{2}.
    \end{aligned}
  \end{equation}
  Minimizing the right-hand side of \eqref{eq: (app-a) newton-square-completion} amounts to minimizing \( \norm{\operatorname{vec}(h) - \bsv^{\star}}_{\fhat_{t}} \) over \( h \) with leaf partition \( \Pi \). The minimum is attained by a tree \( h_{\Pi} \) whose value vector is the projection of \( \bsv^{\star} \) onto the subspace of vectors that are constant on each leaf of \( \Pi \). By orthogonality, this gives
  \[
    \frac{1}{2}
    \norm{\operatorname{vec}(h_{\Pi}) - \bsv^{\star}}_{\fhat_{t}}^{2} -
    \frac{1}{2}
    \norm{\bsv^{\star}}_{\fhat_{t}}^{2} =
    -\frac{1}{2}
    \norm{\operatorname{vec}(h_{\Pi})}_{\fhat_{t}}^{2}.
  \]
  On the other hand, the left-hand side of \eqref{eq: (app-a) newton-square-completion} is the increment in the second-order Taylor polynomial \eqref{eq: (sec-2) second-order-taylor}. As shown in \SECref{sec: (sec-2) problem-algorithm}, its minimum over the leaf values is \( -\Dhat_{t}(\Pi) \). Equating the two minima gives \( \norm{\operatorname{vec}(h_{\Pi})}_{\fhat_{t}}^{2} = 2 \Dhat_{t}(\Pi) \). For the selected leaf partition \( \Pi = \Pihat_{t} \) (recalling that we write \( \Dhat_{t} = \Dhat_{t}(\Pihat_{t}) \)), this gives \( \norm{\operatorname{vec}(\hhat_{t})}_{\fhat_{t}}^{2} = 2 \Dhat_{t} \), proving the second claim.
\end{proofof}

The preceding proof explains the geometric motivation for \ASMref{asm: (sec-3) progress-toward-comparison}. We now give a sufficient condition under which this assumption holds.

\begin{lemma}
  \label{lem: (app-a) sufficient-progress}
  Suppose that, for every \( t = 0, \dots, K - 1 \) and each nonempty leaf \( A \in \Pihat_{t} \), there are constants \( \vartheta_{t}(A) \) and \( \vartheta^{\mathrm{ref}}(A) \) such that \( \fhat_{t}(x) = \vartheta_{t}(A) \) and \( \fref(x) = \vartheta^{\mathrm{ref}}(A) \) for every training observation \( x \in A \). Then
  \[
    2 \Dhat_{t} \ge
    \RiskN{\fhat_{t}} - \RiskN{\fref}
  \]
  for every \( t = 0, \dots, K - 1 \). In particular, \ASMref{asm: (sec-3) progress-toward-comparison} holds with \( \omega = 1 \).
\end{lemma}

\begin{proof}
  Fix \( t \). By \eqref{eq: (sec-2) leaf-derivatives}, each nonempty leaf \( A \in \Pihat_{t} \) satisfies
  \begin{equation}
    \label{eq: (app-a) leafwise-progress}
    \begin{aligned}
      2 \frac{\Ghat_{t}(A)^{2}}{2 \Hhat_{t}(A)}
       & =
      \Dparen*{
        1 + \frac{
          2 \sqrt{\ahat_{t}(A) \bhat_{t}(A)}
        }{
          \ahat_{t}(A) + \bhat_{t}(A)
        }
      }
      \Dparen*{
        \frac{\ahat_{t}(A) + \bhat_{t}(A)}{2} -
        \sqrt{\ahat_{t}(A) \bhat_{t}(A)}
      }
      \\ & \ge
      \frac{\ahat_{t}(A) + \bhat_{t}(A)}{2} -
      \sqrt{\ahat_{t}(A) \bhat_{t}(A)}.
    \end{aligned}
  \end{equation}
  Let \( \vartheta_{t}(A) \) denote the common value of \( \fhat_{t} \) on the training observations in \( A \). By \eqref{eq: (sec-2) leaf-masses},
  \[
    \ahat_{t}(A) =
    \ee^{-\vartheta_{t}(A) / 2}
    \PP[P_{N}]{X \in A},
    \quad
    \bhat_{t}(A) =
    \ee^{\vartheta_{t}(A) / 2}
    \PP[Q_{N}]{X' \in A},
  \]
  so their exponential factors cancel in \( \sqrt{\ahat_{t}(A) \bhat_{t}(A)} \), while \( \sum_{A \in \Pihat_{t}} (\ahat_{t}(A) + \bhat_{t}(A)) / 2 = \RiskN{\fhat_{t}} \). Consequently, summing \eqref{eq: (app-a) leafwise-progress} over \( A \in \Pihat_{t} \) gives
  \begin{equation}
    \label{eq: (app-a) progress-risk-bound}
    2 \Dhat_{t} \ge
    \RiskN{\fhat_{t}} -
    \sum_{A \in \Pihat_{t}}
    \sqrt{
      \PP[P_{N}]{X \in A}
      \PP[Q_{N}]{X' \in A}
    }.
  \end{equation}
  Turning to \( \fref \), let \( \vartheta^{\mathrm{ref}}(A) \) denote its common value on the training observations in each nonempty leaf \( A \). The arithmetic--geometric mean inequality gives
  \begin{equation}
    \label{eq: (app-a) comparison-risk-bound}
    \begin{aligned}
      \RiskN{\fref}
       & =
      \frac{1}{2}
      \sum_{A \in \Pihat_{t}}
      \Dparen*{
        \ee^{-\vartheta^{\mathrm{ref}}(A) / 2}
        \PP[P_{N}]{X \in A} +
        \ee^{\vartheta^{\mathrm{ref}}(A) / 2}
        \PP[Q_{N}]{X' \in A}
      }
      \\ & \ge
      \sum_{A \in \Pihat_{t}}
      \sqrt{
        \PP[P_{N}]{X \in A}
        \PP[Q_{N}]{X' \in A}
      }.
    \end{aligned}
  \end{equation}
  Combining \eqref{eq: (app-a) progress-risk-bound} and \eqref{eq: (app-a) comparison-risk-bound} proves the lemma.
\end{proof}

We now give an example satisfying Assumptions~\ref{asm: (sec-3) density-ratio-tails}--\ref{asm: (sec-3) local-smoothness} and show, using \LEMref{lem: (app-a) sufficient-progress}, that \ASMref{asm: (sec-3) progress-toward-comparison} holds with positive probability.

\noindent \textbf{A concrete example.} Let \( P \) and \( Q \) be supported on \( \family{-1, 1} \), with
\begin{equation}
  \label{eq: (app-a) two-point-progress-example}
  \PP[P]{X = -1} = \PP[Q]{X' = 1} = \frac{3}{4},
  \quad
  \PP[P]{X = 1} = \PP[Q]{X' = -1} = \frac{1}{4}.
\end{equation}
In this setting, Assumptions~\ref{asm: (sec-3) density-ratio-tails} and \ref{asm: (sec-3) input-tails} hold with \( \kratio = 6 \) and \( \kinput = 2 \). Extend the log density ratio to \( \bbR \) by setting \( \fstar(x) = -x \log 3 \). Then \ASMref{asm: (sec-3) local-smoothness} also holds with \( \alpha_{1} = 1 \), \( \zeta = 1 \), and \( H_{0} = 2 \log 3 \).

Let \( \widehat{\calC}_{1} = \family{-2, 0, 2} \) be the set of allowed split points. Take \( u = 2 \) for the cube \( (-u, u] \), and let \( L > 4 \) be the leaf limit. By the construction in \APPref{subsec: (app-a) approximation-proof}, the leaf partition of the comparison tree is
\[
  \PartionApp = \family{
    (-\infty, -2],
    (-2, 0],
    (0, 2],
    (2, \infty)
  }.
\]
Suppose we observe training samples whose empirical frequencies match \eqref{eq: (app-a) two-point-progress-example}. This event has positive probability. With a clipping threshold \( B_{N} \ge \log 3 \), the corresponding values of \( \fref \) on these four leaves are \( 0 \), \( \log 3 \), \( -\log 3 \), and \( 0 \), respectively. Furthermore, it can be shown that, at every iteration, Algorithm~\ref{alg: (sec-2) training} makes exactly one split at zero. Consequently, for \( 0 < \nu \le 1 \) and \( t = 0, 1, \dots \), we have
\[
  \Pihat_{t} =
  \family{(-\infty, 0], (0, \infty)},
  \quad
  \fhat_{t}(x) =
  \left\{
  \begin{array}{ll}
    \vartheta_{t},  & x \in (-\infty, 0]; \\
    -\vartheta_{t}, & x \in (0, \infty),
  \end{array}
  \right.
\]
where \( \vartheta_{0} = 0 \) and
\[
  \vartheta_{t + 1} =
  \vartheta_{t} +
  2 \nu \tanh \Dparen*{
    \frac{\log 3 - \vartheta_{t}}{2}
  }.
\]
All training observations take values in \( \family{-1, 1} \), and each leaf of \( \Pihat_{t} \) contains only one of these two values. Hence both \( \fhat_{t} \) and \( \fref \) are constant on the training observations within each leaf. By \LEMref{lem: (app-a) sufficient-progress}, \ASMref{asm: (sec-3) progress-toward-comparison} holds with \( \omega = 1 \).

\subsection{Proofs of the Unified Bounds and Model Selection}
\label{subsec: (app-a) unified-validation-proofs}

In this subsection, we prove \THMref{thm: (sec-3) unified-finite-sample}, \CORref{cor: (sec-3) explicit-rate}, and \THMref{thm: (sec-3) validation-selection}.

\begin{proofof}{\THMref{thm: (sec-3) unified-finite-sample}}
  Since \( \fapp \) is one of the trees over which \( \fref \) minimizes the empirical risk, the definition \eqref{eq: (sec-3) optimization-error-definition} of \( \Opterror \) gives
  \[
    \RiskN{\fhat_{K, B_{N}}} \le
    \RiskN{\fref} + \Opterror \le
    \RiskN{\fapp} + \Opterror.
  \]
  We take \( \delta / 2 \) as the failure probability in each of \THMref{thm: (sec-3) estimation-error} and \THMref{thm: (sec-3) optimization-error}, so their conclusions hold simultaneously with probability at least \( 1 - \delta \). The bound \eqref{eq: (sec-3) estimation-comparison-bound} then applies with \( \fhat = \fhat_{K, B_{N}} \), \( f = \fapp \), and \( \varepsilon = \Opterror \). Substituting the bounds on \( \ExRisk{\fapp} \) and \( \Opterror \) from \THMref{thm: (sec-3) approximation-error} and \THMref{thm: (sec-3) optimization-error}, respectively, and collecting terms proves \eqref{eq: (sec-3) explicit-main-bound} in \THMref{thm: (sec-3) unified-finite-sample}. The bound holds for every trajectory satisfying \ASMref{asm: (sec-3) progress-toward-comparison}. Without this assumption, the same argument gives the bound with \( E_{3} \) replaced by \( \Opterror \), as stated in \THMref{thm: (sec-3) unified-finite-sample}.
\end{proofof}

\begin{proofof}{\CORref{cor: (sec-3) explicit-rate}}
  By \THMref{thm: (sec-3) unified-finite-sample}, it suffices to bound the four terms \( E_{1} \), \( E_{2} \), \( E_{3} \), and \( E_{4} \) in its conclusion. We choose
  \[
    \begin{aligned}
      \ee^{B_{N}}
       & = 8 \kratio \log N,
       & u_{N}
       & = \kinput \log N,
      \\
      K
       & = \left\lceil
      \frac{2 \log N}{
        (2 \nu - \nu^{2}) \omega
      }
      \right\rceil,
       & L
       & = \left\lceil
      \Dparen*{
        N (\log N)^{2 \zeta - 3}
      }^{1 / (2 \alphabar + 1)}
      \right\rceil,
    \end{aligned}
  \]
  and take \( N \) large enough that these choices satisfy the conditions of \THMref{thm: (sec-3) unified-finite-sample}. Our choices of \( K \), \( B_{N} \), and \( u_{N} \) then ensure
  \[
    E_{3} =
    \exp \Dparen*{
      -\frac{2 \nu - \nu^{2}}{2} \omega K
    }
    \le N^{-1},
    \quad
    E_{4} =
    \exp \Dparen*{
      -\frac{\ee^{B_{N}}}{8 \kratio}
    } +
    \exp \Dparen*{
      -\frac{u_{N}}{\kinput}
    }
    \le 2 N^{-1}.
  \]
  It remains to bound \( E_{1} \) and \( E_{2} \). Since \( \ee^{B_{N}} \), \( K \), \( 1 + u_{N} \), and \( \log(d K L) \) are all at most \( C \log N \), we have
  \[
    E_{1} \le
    C \Dparen*{
      \frac{L (\log N)^{7 / 2}}{N} +
      \frac{
        (\log N)^{1 / 2} \log(8 / \delta)
      }{N}
    },
    \quad
    E_{2} \le
    C (\log N)^{2 \zeta + 1 / 2}
    L^{-2 \alphabar}.
  \]
  For \( E_{2} \), condition \eqref{eq: (sec-3) grid-condition} allows the terms involving \( \Resolution \) to be absorbed into the term involving \( L \), as shown in \THMref{thm: (sec-3) approximation-error}. Our choice of \( L \) balances the first summand in the bound on \( E_{1} \) with the bound on \( E_{2} \), so that both are at most
  \[
    C N^{-\frac{2 \alphabar}{2 \alphabar + 1}}
    (\log N)^{
        \frac{
          7 \alphabar + 2 \zeta + 1 / 2
        }{2 \alphabar + 1}
      }.
  \]
  This is the first term of \eqref{eq: (sec-3) convergence-rate} in \CORref{cor: (sec-3) explicit-rate}. The second summand of \( E_{1} \) is its second term, which also absorbs \( E_{3} \) and \( E_{4} \).  Since \THMref{thm: (sec-3) unified-finite-sample} holds with probability at least \( 1 - \delta \), so does \eqref{eq: (sec-3) convergence-rate}. The proof is complete.
\end{proofof}

\begin{proofof}{\THMref{thm: (sec-3) validation-selection}}
  Condition on the training samples. We reuse the definition \eqref{eq: (app-a) normalized-loss-difference} of \( g_{f} \) from the proof of \THMref{thm: (sec-3) estimation-error}. Consider the class
  \[
    \calA = \family{
      g_{\fhat^{\langle 1 \rangle}},
      \dots,
      g_{\fhat^{\langle M \rangle}}
    } \cup \family{0}.
  \]
  The functions in \( \calA \) satisfy the conditions of \LEMref{lem: (app-a) peeled-relative-deviation}. Moreover, since \( \abs{\calA} \le M + 1 \), we have
  \[
    \log \calN(\eta, \calA, \scrL^{1}(S)) \le
    \log \abs{\calA} \le
    C \log M.
  \]
  We group the validation observations into \( N_{v} \) independent pairs with distribution \( W = P \otimes Q \). Applying \LEMref{lem: (app-a) peeled-relative-deviation} to \( \calA \) with this distribution, we obtain
  \begin{equation}
    \label{eq: (app-a) validation-relative-deviation}
    \begin{aligned}
       & \eqspace
      \abs{
        \Dparen*{
          \RiskNv{\fhat^{\langle m \rangle}} -
          \RiskNv{\fstar_{B_{N}}}
        } -
        \Dparen*{
          \Risk{\fhat^{\langle m \rangle}} -
          \Risk{\fstar_{B_{N}}}
        }
      }
      \\ & \le
      \frac{1}{4} \Dparen*{
        \Risk{\fhat^{\langle m \rangle}} -
        \Risk{\fstar_{B_{N}}}
      } +
      C \ee^{B_{N} / 2} \frac{
        (\log M) \log N_{v} +
        \log(2 / \delta)
      }{N_{v}},
    \end{aligned}
  \end{equation}
  simultaneously for all \( 1 \le m \le M \), with probability at least \( 1 - \delta \). Let \( \mhat \) be an index minimizing the validation risk \( \widehat{\calR}_{v} \). Expanding the absolute value in \eqref{eq: (app-a) validation-relative-deviation} and applying the resulting lower bound to \( \mhat \) and upper bound to any \( m \), we obtain
  \[
    \begin{aligned}
       & \eqspace
      \frac{3}{4} \Dparen*{
        \Risk{\fhat^{\langle \mhat \rangle}} -
        \Risk{\fstar_{B_{N}}}
      }
      \\ & \le
      \frac{5}{4} \Dparen*{
        \Risk{\fhat^{\langle m \rangle}} -
        \Risk{\fstar_{B_{N}}}
      } +
      C \ee^{B_{N} / 2} \frac{
        (\log M) \log N_{v} +
        \log(2 / \delta)
      }{N_{v}}.
    \end{aligned}
  \]
  Minimizing over \( m \) and rearranging proves \eqref{eq: (sec-3) validation-bound} in \THMref{thm: (sec-3) validation-selection}.
\end{proofof}

\bibliography{reference}

\end{document}